\pdfoutput=1

\documentclass[11pt]{article}

\usepackage[final]{coling}

\usepackage{times}
\usepackage{latexsym}

\usepackage[T1]{fontenc}

\usepackage[utf8]{inputenc}
\usepackage{hyperref}
\usepackage{microtype}

\usepackage{inconsolata}

\usepackage{graphicx}

\usepackage{enumitem}  
\usepackage{amsmath}   
\usepackage{amsfonts}  
\usepackage{booktabs}  
\usepackage{multirow}  

\usepackage{booktabs}
\usepackage{multirow}
\usepackage{graphicx}
\usepackage{caption}
\usepackage{subcaption}
\usepackage{amsfonts}
\usepackage{algorithm}
\usepackage{algorithmic}
\usepackage{amssymb}
\usepackage{pifont}
\usepackage{xcolor}
\usepackage{bm}

\newcommand{\mB}[0]{\mathbf{B}}

\newcommand{\mX}[0]{\mathbf{X}}

\DeclareMathOperator{\norm}{Norm}
\DeclareMathOperator{\sigmoid}{Sigmoid}
\DeclareMathOperator{\sparsity}{Sparsity}

\def\1{\bm{1}}

\def\vx{{\bm{x}}}

\newcommand{\ours}[0]{\textsc{CFSP}}

\title{\ours{}:
An Efficient Structured Pruning Framework for LLMs with Coarse-to-Fine Activation Information 
}

\author{
\textbf{
    Yuxin Wang$^{1}$\footnotemark[1],
    Minghua Ma$^{1}$\footnotemark[1],
    Zekun Wang$^{1}$\thanks{Equal Contribution.}\thanks{Corresponding Author.},
    Jingchang Chen$^{1}$, 
    Huiming Fan$^{1}$}, \\
\textbf{
    Liping Shan$^{2}$,
    Qing Yang$^2$,
    Dongliang Xu$^2$,
    Ming Liu$^{1}$\footnotemark[2],
    Bing Qin$^{1}$} \\
    $^{1}$Harbin Institute of Technology, Harbin, China \\
    $^{2}$Du Xiaoman Science Technology Co., Ltd, Beijing, China \\
 \texttt{\{yuxinwang,mhma,zkwang,jcchen,mliu,qinb\}@ir.hit.edu.cn} \quad
}

\begin{document}
\maketitle

\begin{abstract}
The colossal parameters and computational overhead of Large Language Models (LLMs) challenge their real-world applications.
Network pruning, which targets unstructured or structured sparsity by removing redundant parameters, has recently been explored for LLM acceleration.
Existing LLM pruning works focus on unstructured pruning, which typically requires special hardware support for a practical speed-up. 
In contrast, structured pruning can reduce latency on general devices.
However, it remains a challenge to perform structured pruning efficiently and maintain performance, especially at high sparsity ratios.
To this end, we introduce an efficient structured pruning framework named \ours{}, which leverages both \textbf{C}oarse (interblock) and \textbf{F}ine-grained (intrablock) activation information as an importance criterion to guide pruning.
The pruning is highly efficient, as it only requires one forward pass to compute feature activations.
Specifically, we first allocate the sparsity budget across blocks based on their importance and then retain important weights within each block.
In addition, we introduce a recovery fine-tuning strategy that adaptively allocates training overhead based on coarse-grained importance to further improve performance.
Experimental results demonstrate that \ours{} outperforms existing methods on diverse models across various sparsity budgets. Our code will be available at \url{https://github.com/wyxscir/CFSP}.
\end{abstract}

\section{Introduction}
Although scaling up Large Language Models (LLMs) brings remarkable performance~\citep{gpt3,openai2023gpt4,team2023gemini,llama3,deepseekai2024deepseekv2,qwen2}, increasing parameters brings more computations and memory consumption, posing a significant challenge of deploying in practical applications.
To address this, various model compression methods for LLMs are proposed~\citep{dettmers2022LLMint8, frantar2022gptq, lin2024awq, minitron2024}.
Existing LLM pruning work~\citep{sparsegpt,wanda,xu2024besa,zhang2024plugandplay} focuses mainly on unstructured or semi-structured sparsity.
However, these paradigms require specific hardware to achieve practical acceleration.

In contrast, \textit{structured pruning}, which imposes structured sparsity by removing groups of consecutive parameters~\citep{louizos2017learning, wang2020structured, xia2022structured}, is more hardware-friendly on general devices.
However, there are some challenges involved in existing structured pruning methods for LLMs:
(1) 
They typically introduce learnable masks to search~\citep{ShearedLLaMA,bonsai} or utilize gradients to guide pruning~\citep{ma2023llmPruner,loraprune}.
Unfortunately, they require significant computational overhead, especially for large-scale (\textit{e.g.}, 70B) models.
(2)
It is also worth noting that they usually assign a uniform sparsity budget per block, which is suboptimal since LLM blocks have different significance in the representation functionality~\citep{gromov2024the}.
Moreover, they usually involve a recovery fine-tuning with Low-Rank Adapter (LoRA)~\citep{hu2022lora} to enhance pruned models, which also introduce training overhead and overlook the varying importance of blocks.

\begin{figure*}[t!]
\centering
\includegraphics[width=1.0\textwidth]{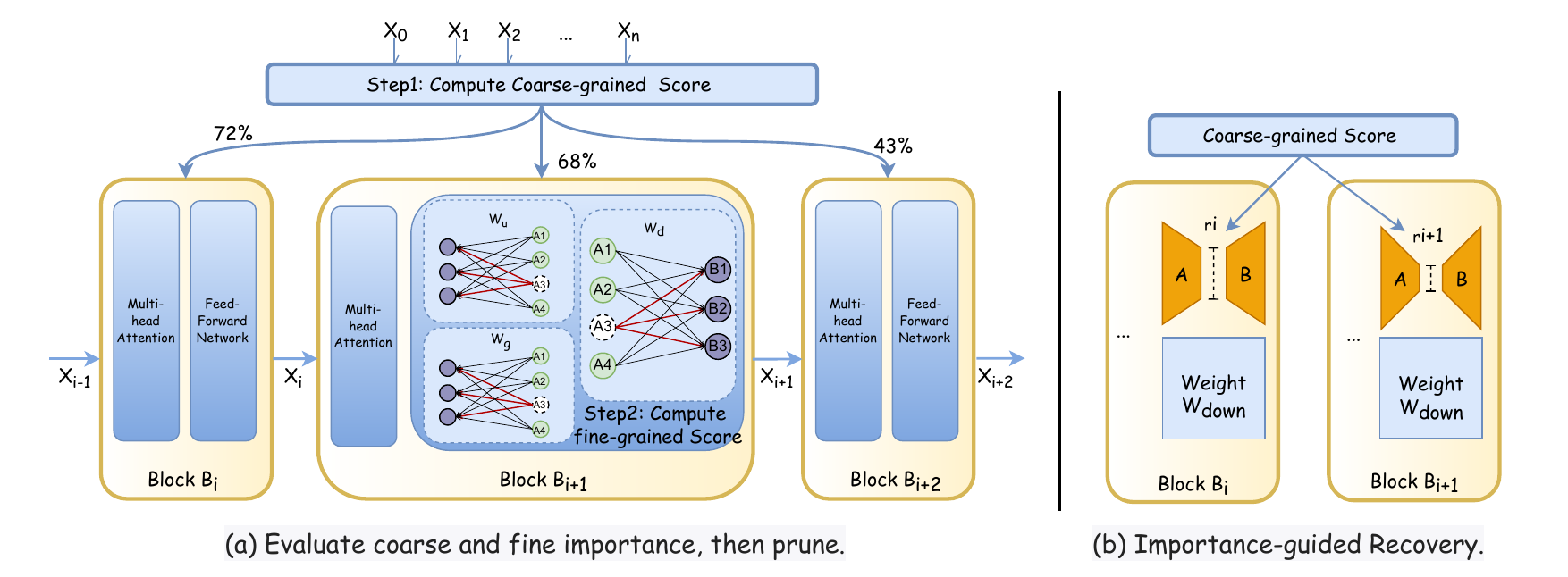}
\caption{
Illustration of our proposed \ours{} framework.
(a) Pruning with coarse (interblock) to fine (intrablock) activation information guidance.
(b) Recovery fine-tuning with importance-guided allocation, where the rank sizes of each component are determined by coarse-grained importance.}
\label{fig:main_pic}
\end{figure*}

To this end, we propose \ours{} (shown in Figure~\ref{fig:main_pic}), an efficient structural pruning framework for LLM that takes advantage of coarse to fine-grained activation information to guide pruning.
Specifically, we employ activation as the importance criterion, which is calculated for blocks (coarse-grained) and the weights within each block (fine-grained) in a single forward pass.
For each block, we measure its saliency of transformations on the basis of the angular distance of the input and output representations.
Then, we utilize this criterion as coarse-grained importance to assign the sparsity budget.
Finally, for weights within each block, we take the product of their relative activations and weights as a fine-grained criterion to remove redundant parts.
Since existing work typically performs recovery fine-tuning with LoRA to further improve performance, we propose a more efficient recovery method that leverages coarse-grained importance to adaptively allocate additional trainable parameters: the pruned models can achieve comparable performance while utilizing less recovery data.

Our contributions can be summarized as follows:
\begin{itemize}[itemsep=2pt,topsep=2pt,parsep=0pt,leftmargin=*]
  \item We propose an efficient coarse-to-fine importance criterion for identifying redundant structures for pruning, which takes only a few minutes\footnote{Details of time cost are shown in Table~\ref{tab:time} in Appendix.} to complete on various models.
  \item We introduce an efficient recovery fine-tuning method that adaptively assigns additional trainable parameters based on the coarse-grained importance score.
  \item Extensive experimental results indicate that \ours{} surpasses existing methods across various models at different sparsity levels, demonstrating promising performance on challenging tasks even at high sparsity levels.
\end{itemize}

\section{Methodology}
\label{sec:methodology}
The overview of \ours{} framework is shown in Figure~\ref{fig:main_pic}.
We first introduce our preliminary analysis in Section \ref{subsec:Preliminaries}, then give details of our pruning criterion and procedure in Section \ref{subsec:tap_importance}.
Finally, we introduce the proposed importance-guidance recovery strategy in Section \ref{subsec:recover}.

\subsection{Preliminaries}
\label{subsec:Preliminaries}
\label{para:Pre_experiment}
\begin{figure}[t!]
\centering

\includegraphics[width=0.9\linewidth]{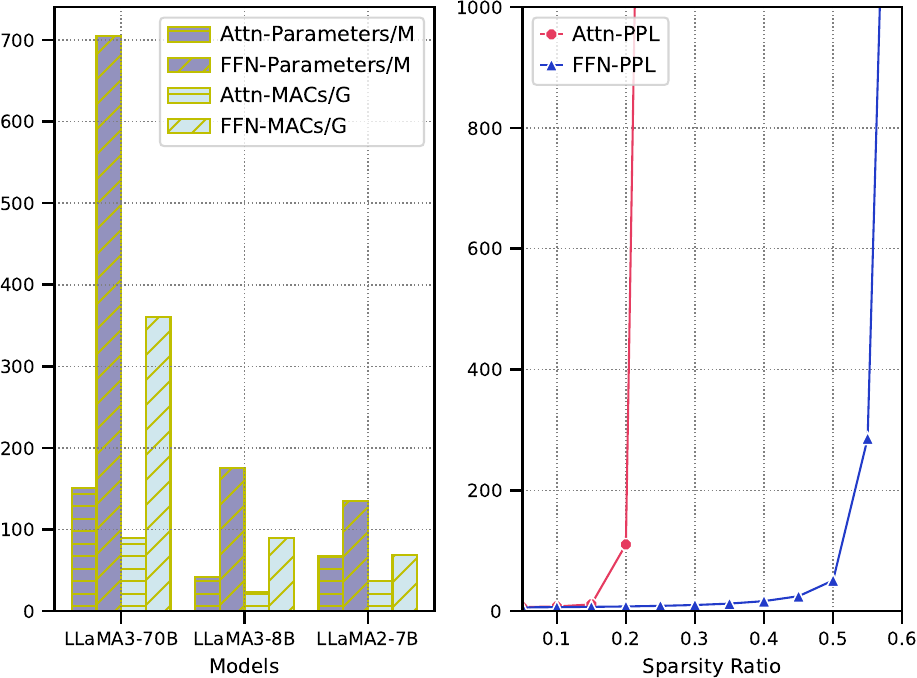}
\caption{
Preliminary analysis. (Left): Parameter size and MACs of modules. (Right): Sensitivity of pruning each module on LLaMA2-7B.}\label{fig:preliminary}
\end{figure}

The Transformer block~\citep{vaswani2017attention} consists of multi-head attention (MHA) and feedforward network (FFN).
We analyze the computational overhead and the sparsity of them.
As shown in Figure~\ref{fig:preliminary}, the parameter size and MAC of FFN are significantly larger than those of MHA.
In addition, we observe that pruning MHA leads to a significant performance drop with only 10\% sparsity, while pruning FFN has a more stable performance even with 50\% sparsity, showing that the FFN module has a higher structural sparsity~\citep{zhang2022moefication} and is more friendly to structured pruning~\citep{gunter2024apple}.
Thus, in this work, we focus on pruning the intermediate dimension of FFN.

\subsection{\ours{} Framework}
\label{subsec:tap_importance}

\ours{} takes activations as an importance criterion to identify redundant parts of LLMs for the following reasons:
(1) Activations can be obtained with a single forward pass, resulting in significantly lower overhead compared to other metrics.
(2) As pointed out in previous studies~\citep{wanda,lin2024awq}, parameter weights corresponding to larger activation magnitudes are more salient since they process more important features.

The feature activations are calculated on a small number (\textit{e.g.}, 128) of calibration samples.
We further incorporate coarse- and fine-grained importance for sparsity allocation and weight pruning.

\paragraph{Coarse-grained Importance}
Existing pruning work usually assigns the same sparsity to each block, but it is suboptimal for sparsity allocation.
In fact, the blocks perform different functions and their importance varies significantly~\citep{ UnreasonableIneffectiveness}.
Due to the residual structure, the effect of each block can be viewed as a transformation of the input representations.

Thus, we measure the coarse-grained importance of blocks $S_g$ through the saliency of transformation of feature activations during the forward process. Specifically, the $S_g$ of $l$-th block $\mB^{\ell}$ is calculated as:
\begin{align}
  S_g(\mB^{\ell}) & = \sum_{i=0} D(\vx^{\ell}_i, \vx^{\ell+1}_i)\\\label{eq:arccos-sim}
  D({\vx}^{\ell}_i, {\vx}^{\ell+1}_i) & = \frac 1\pi \arccos(\frac{{\vx}^{\ell}_i\cdot {\vx}^{\ell+1}_i}{\Vert{\vx}^{\ell}_i\Vert \Vert{\vx}^{\ell+1}_{i}\Vert })\
\end{align}
where $\vx^{\ell}$ and $\vx^{\ell+1}$ represent the input and output activation states of the $\ell$-th block.
$D(\cdot)$ can be various distance measurements of two representations.
Here we select the angular distance because it performs better than the others in our experiments.
We then normalize $\mB^{\ell}$ with the $\sigmoid$ as:
\begin{align}\label{eq:22}
    \norm(S_g(\mB^{\ell})) & = \sigmoid(S_g(\mB^{\ell}) - \overline{S}) \\
    \sigmoid(x) & = \frac{1}{1 + e^{- \alpha \cdot x}}
\end{align}
where $\overline{S}$ is the average importance score of all blocks. The function $\sigmoid$ is introduced to process the scores non-linearly, which can make the distinction between blocks more significant, and $\alpha$ controls the intensity of significance.
Finally, we assign the sparsity budgets across blocks based on the normalized importance scores as:
\begin{equation}\label{eq:final_global}
  \sparsity(\mB^{\ell}) = \frac {\text{Norm}(S_g(\mB^{\ell})) \cdot \gamma \cdot n}{\sum_{\ell=1}^{n} \norm(S_g(\mB^{\ell}))}
\end{equation}
where $n$ is the number of blocks and $\gamma$ represents the whole sparsity budget of the model.

\begin{figure}[t!]
\centering
\includegraphics[width=0.45\textwidth]{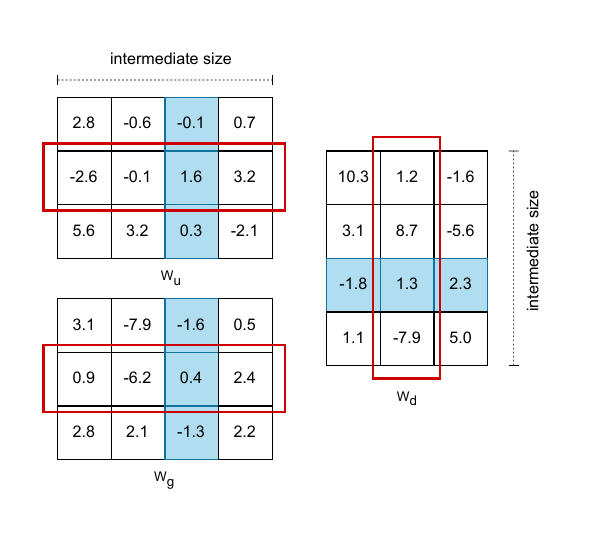}
\caption{The structural dependencies of FFN in LLaMA3. 
The blue part corresponds to the minimum unit of structured pruning. The red box represents the relative size of a matrix element in its row or column.}
\label{fig:ffn_dependency}
\end{figure}

\paragraph{Dimension Adjustment} 

Equation~\ref{eq:final_global} can assign irregularly shaped weight matrices that do not satisfy the multiples of 64 or 128, thus destroying the parallelism of the tensor cores on the GPU.
To this end, we introduce a simple adjustment during pruning to adjust the final dimensions of the pruned blocks to multiples of 128.
For the $l$-th block $\mB^{l}$, the final dimension $dim_{f}^l$ are computed as:
\begin{align}\label{eq:acclerate}
  dim_{f}^{l} & = \left( \left\lfloor \frac{dim_{o} \times  \sparsity(\mB^{i}) + 64}{128} \right\rfloor \right) \times 128 
\end{align}
where $dim_{o}$ is the intermediate dimensions of original dense model. 
We present ablation results in Section~\ref{sec:analysis} that demonstrate that the adjustment of dimensions significantly accelerates inference speed on GPUs. Notably, this enhancement is achieved with only a minimal increase in parameters and no detrimental impact on performance.

\paragraph{Fine-grained Importance}
After assigning the proper sparsity to each block, we then identify the importance of pruning units (intermediate dimensions) within the block.
Figure~\ref{fig:ffn_dependency} shows the structural dependencies of three matrices used in FFN ($W_u$, $W_g$ and $W_d$): removing the intermediate dimensions is equivalent to subtracting the corresponding columns of $W_u, W_g$ and the corresponding rows of $W_d$. 
The weight matrix represents the connections between neurons, where each row or column of weights influences the same neuron, implying that the fine-grained importance of the weights is also related to their respective row or column.
For the $i$-th intermediate dimension, we utilize the activation of $\mX_d$ of $W_d$ and all weight matrices to calculate fine-grained importance score $S_l^i$:
\begin{align}
  S_l^i & =  F_l^i \cdot\left\Vert{\mX}_{d}^{i}\right\Vert \\\label{eq:24}
  F_l^i & =  \sum_{j} \left(\frac {W^{ij}_{d}\cdot \left\Vert{\mX}^{i}_{d}\right\Vert } {W^{*j}_{d}\cdot\left\Vert{\mX}^{*}_{d}\right\Vert} 
  + \frac {W^{ij}_{u}}{W^{i*}_{u}}
  + \frac {W^{ij}_{g}}{W^{i*}_{g}}\right)
\end{align}
where $\left\Vert \cdot \right\Vert$ is L2 normalization. As shown in Equation~\ref{eq:24},
the pruning structure $F$ consists of three matrices determined by cumulative activation of the matrix $W_d$. 
The matrices $W_u$ and $W_g$ use a relative weight measurement, where the magnitude of the weight is proportional to the sum of the row in the matrices.
An unique aspect is the matrix $W_d$ in $F$, which quantifies the ratio between the weight activation magnitudes and the sum of column activation magnitudes of $W_d$, as shown in the first term of Equation~\ref{eq:24}.

\subsection{Importance-guided Recovery Fine-tuning}
\label{subsec:recover}
In addition to the single-shot pruning scenario, we also explore the integration of recovery fine-tuning to further enhance performance at high sparsity.
Our recovery setting follows \citet{ma2023llmPruner} to fine-tune with LoRA~\citep{hu2022lora}.
Unlike the original LoRA, we propose an importance-guided method that adaptively assigns additional trainable parameters across different blocks. 
Specifically, for the $l$-th block, the rank $r^l$ of LoRA is determined based on the coarse-grained importance scores computed during pruning:
\begin{gather}
r^{l} =  \frac {\norm(S_g(\mB^{\ell})) \cdot \bar r \cdot n}{\sum_{\ell=1}^{n} \norm(S_g(\mB^{\ell}))}
\label{eq:ri}
\end{gather} 
where $\bar r$ is the averaged rank allocated budget. 
In our experiments, we find that our recovery method is more efficient, achieving comparable performance while requiring less training data.

\paragraph{Recovery Data} 
We explore various datasets for recovery fine-tuning.
We find that the quality and diversity of knowledge in the data are critical for recovery performance, especially on challenging tasks (\textit{e.g.}, MMLU). 
The details of datasets and results can be found in Appendices \ref{app:methods_details} and \ref{app:more_analysis}.

\begin{table*}[t!]
  \centering
  \setlength{\tabcolsep}{4.5pt}
  \resizebox{1.0\linewidth}{!}{
  \begin{tabular}{l|l|cccccccc|c}
  \toprule
  Sparsity & Method & WinoGrande & PIQA & HellaSwag & OBQA & ARC-e & ARC-c & MMLU & FreebaseQA & Average \\
  \midrule
    0\% & LLaMA3-8B &  72.93 & 80.96  & 79.17 & 45.00 & 77.90 & 53.16 &  62.09 & 72.62 & 67.98\\
    \midrule 
    \multirow{5}{*}{20\%}  &\multicolumn{8}{l}{\textit{w/o recovery}}  \\
    & Magnitude-SP & 50.99 & 51.31 & 26.18 & 30.20 & 25.43& 25.76 & 23.71 &0.52 & 29.26 \\ 
    & Wanda-SP & 67.56 & 75.41  & 65.99 & \bf 42.00& 65.40 & 41.38& 46.20 & \textbf{39.11} & 55.38 \\
    & FLAP & 65.67 & 74.65 & 62.41 & 40.20 & 61.36 & 35.15 & 41.39 & 34.58 & 51.93\\
    & \ours{} (ours) & \textbf{70.32} & \textbf{77.64} & \textbf{72.74} & 41.20 & \textbf{68.10} & \textbf{43.86} & \textbf{56.43} & 38.59 & \textbf{58.61} \\
    \midrule
    \multirow{8}{*}{50\%} &\multicolumn{8}{l}{\textit{w/o recovery}}  \\
    & Magnitude-SP & 51.14 & 50.27 & 26.47 & 29.40 & 25.08 & 26.49 & 23.08 & 0.52 & 29.06 \\   
    & Wanda-SP & 59.51 & 63.98 & 45.71 & \bf 33.00 & 44.95 & 27.99 & 27.08 & 6.15 & 38.54\\
    & FLAP & 58.80 & 62.35& 41.89 & 31.00 & 40.28 & 26.11 & 24.03 & 4.55 & 36.12 \\
    &\ours{} (ours) & \textbf{62.04} & \textbf{66.76}  & \textbf{49.96} & 31.80 & \bf 48.74 & \bf 30.89 & \textbf{32.39} & \textbf{10.83} & \textbf{41.67} \\
     \cmidrule(lr){2-11} 
    & \multicolumn{8}{l}{\textit{w/ recovery}} \\
    & Wanda-SP & 61.48 & 70.89 & 60.20& \bf 37.60 & 60.86 &36.43  & 35.54 & 11.89& 46.86 \\
    &\ours{} (ours) &  \textbf{65.51} & \textbf{72.03}  & \textbf{61.45} & 36.20 & \textbf{62.37} & \textbf{37.54} & \textbf{40.37} & \textbf{18.32} & \textbf{49.22} \\
  \bottomrule
  \end{tabular}
      }
  \caption{Zero-shot performance of pruned models on LLaMA3-8B under 20\% and 50\% sparsity.
  For 50\% sparsity, we also show the results after recovery fine-tuning.
  \textbf{Bold} indicates the best results under the same setting.}
  \label{tab:zs_llama3_8b}
\end{table*}

\begin{table*}[t]
\centering
\setlength{\tabcolsep}{4.5pt}
\resizebox{1.0\linewidth}{!}{
\begin{tabular}{l|l|cccccccc|c}
\toprule
Sparsity &  Method & WinoGrande & PIQA &  HellaSwag & OBQA & ARC-e & ARC-c & MMLU & FreebaseQA & Average\\
\midrule
  0\%  &LLaMA3-70B &  80.35 & 84.71 & 84.93 & 48.60 &  85.90 & 64.16 & 75.36 & 81.53 & 75.69\\
  \midrule
      \multirow{5}{*}{20\%} & \multicolumn{8}{l}{\textit{w/o recovery}} \\
    & Magnitude-SP & 51.93 & 58.38 &  32.69 & 29.40 & 32.41 & 27.73 & 24.44 & 0.52 & 32.19 \\
  & Wanda-SP & 77.19 & 82.92 &  82.50 & \bf 49.20 & 81.65 & 58.28 & 66.74 & 79.65 & 72.27 \\
  & FLAP & 77.51 & 82.48 & 80.41 & 47.40 &  78.49 & 55.12 & 65.88 & 79.02 & 70.79 \\
  & \ours{} (ours) & \bf 80.66 & \bf 83.51 &   \bf 83.97 &46.40 & \bf 83.46 & \bf 61.43 & \bf 73.04 & \bf 80.18 & \bf 74.08 \\
  \midrule
\multirow{8}{*}{50\%} & \multicolumn{8}{l}{\textit{w/o recovery}}\\
  & Magnitude-SP & 51.22 & 52.72 &  27.04 & 30.00 & 25.46 & 26.62 & 23.52 & 0.55 & 29.64 \\
  & Wanda-SP & 73.95 & 76.44 &  73.80 & \bf 44.00 & 66.79 & 43.94 & 54.91 & 42.26 & 59.51 \\
  & FLAP & 72.85 & 76.82  & 68.05 & 42.80 & 66.54 & 45.05 & 53.90 & 38.41 & 58.05 \\
  & \ours{} (ours) & \bf 75.06 & \bf 78.89 &  \bf 75.95 & 43.60 & \bf 71.34 & \bf{46.67} & \bf 59.74 & \bf 46.42 & \bf 62.20 \\
       \cmidrule(lr){2-11} 
    & \multicolumn{8}{l}{\textit{w/ recovery}} \\
  & Wanda-SP & 76.33 & 80.02 & 79.73 & \bf 47.20 & 73.10 & 47.26 & 59.98 & 48.20 & 63.98 \\
  & \ours{} (ours) &  \bf{78.30} & \bf{81.01}  &  \bf{80.18} &45.20 & \bf{76.65} & \bf{51.54} & \bf{65.52} & \bf{54.77} & \bf 66.65 \\
\bottomrule
\end{tabular}
}
\caption{Zero-shot performance of pruned models on LLaMA3-70B under 20\% and 50\% sparsity. For 50\% sparsity, we also show the results after recovery fine-tuning.   
\textbf{Bold} indicates the best results under the same setting. }
\label{tab:zs_llama3_70b}
\end{table*}

\begin{table*}[t!]
  \centering
  \setlength{\tabcolsep}{4.5pt}
  \resizebox{1.0\linewidth}{!}{
  \begin{tabular}{l|l|cccccccc|c}
  \toprule
  Sparsity & Method &  WinoGrande & PIQA
 &  OBQA & HellaSwag & ARC-e & ARC-c & MMLU & FreebaseQA & Average \\
  \midrule
     0\% & LLaMA2-13B & 72.22 & 80.52 & 45.20 & 79.38 & 77.48 & 49.06& 50.51 & 67.57  &  65.24 \\
    \midrule
    \multirow{5}{*}{20\%} & \multicolumn{8}{l}{\textit{w/o recovery}} \\
     & Magnitude-SP &  49.96 & 60.01 & 25.60 &  39.89 & 42.93 &  29.86 & 25.51 & 0.65 & 34.30 \\
      &Wanda-SP & 70.01 & \bf 78.45 & 43.00 & 73.87 & 72.56 & 44.28 & 41.70 & 40.69 & 58.07 \\
     & FLAP & 68.27 & 77.58 &  41.40 &  72.58 & 67.47 &  42.58  & 41.15 & 28.63  & 54.96 \\
       & \ours{} (ours) & \bf 71.75 & 78.29 & \bf 43.60 & \bf 75.76 & \bf 73.48 &  \bf 47.27 & \bf 46.99 & \bf 54.65 & \bf 61.47 \\
    \midrule
    \multirow{8}{*}{50$\%$} & \multicolumn{8}{l}{\textit{w/o recovery}}\\
    & Magnitude-SP &  50.75 & 50.16 & 24.20 & 26.17 & 27.19 &  25.85 & 25.20 & 0.53  & 28.76 \\
      &  Wanda-SP &  64.80 &  71.76 &  38.00 &  57.36 & 59.09 & 37.80 & 27.00 & 3.20  & 44.87  \\ 
    &  FLAP & 60.54 & 68.50 & 36.60 & 53.95 & 48.78 & 30.97 & 23.00 & 1.50  & 40.48 \\ 
      & \ours{} (ours) & \bf 64.17 & \bf 71.98 &  \bf 39.40 & \bf 60.28 & \bf 62.33 & \bf 38.05 & \bf 28.24 & \bf 3.65  & \bf 46.01 \\ 
      \cmidrule(lr){2-11} 
    & \multicolumn{8}{l}{\textit{w/ recovery}} \\
     & Wanda-SP &    66.85 & 74.37 & \bf 40.60 & 67.15 & 68.31 & \bf 41.04 & 35.24 & 25.35  & 52.36 \\ 
     & \ours{} (ours) & \bf 67.17 & \bf 74.88 & \bf 40.60 &  \bf 68.60 & \bf 69.23 & 40.87 & \bf36.41 & \bf25.78  & \bf 52.94 \\ 
  \bottomrule
  \end{tabular}
      }
  \caption{Zero-shot performance of pruned models on LLaMA2-13B under 20\% and 50\% sparsity.
  For 50\% sparsity, we also show the results after recovery fine-tuning.
  \textbf{Bold} indicates the best results under the same setting.}
  \label{tab:zs_llama2_13b}
\end{table*}

\section{Experiments}
\label{sec:Experiments}
\subsection{Experimental Setup}
In this work, we target to prune intermediate dimensions of FFN in LLM and conduct experiments primarily on widely used LLM models: LLaMA3-\{8B,70B\}~\citep{llama3} and a middle-size LLaMA2-13B~\citep{llama2}. 
We also conduct experiments on the latest LLaMA3.1-8B~\citep{llama3.1} and more models from LLaMA family in Appendix~\ref{app:more_llama_results}. 
\paragraph{Evaluation Benchmarks}
Following previous work~\citep{wanda, flap}, we evaluate the zero-shot performance of models across 5 well-known tasks: WinoGrande~\citep{WinoGrande}, PIQA~\citep{PIQA}, HellaSwag~\citep{HellaSwag}, ARC-easy and ARC-challenge~\citep{arc}.
Since \citet{llmkick} shows that LLM pruning methods tend to have significantly degraded performance on knowledge-intensive tasks, we also include two challenging QA tasks for zero-shot evaluation: MMLU~\citep{mmlu} and FreebaseQA~\citep{freebaseqa}, which focus on factual knowledge.
For language modeling performance, we evaluate models on WikiText2~\citep{wikitext2}.
Following previous work, we use the LM-Evaluation-Harness~\citep{eval-harness} and LLM-Kick~\citep{llmkick} with default hyperparameters for the corresponding tasks.
More details of the evaluation are shown in Appendix~\ref{app:benchmarks}

\paragraph{Implementation Details}
For the pruning stage, the calibration data are randomly selected from the WikiText2~\citep{wikitext2} training set.
Unless otherwise stated, the calibration set consists of 128 samples and each has approximately 1024 tokens following \citet{wanda}. 
In the recovery fine-tuning stage, pruned models are trained on 0.1B tokens from the FineWeb-Edu~\citep{lozhkov2024fineweb-edu} dataset with the next-token prediction loss.
We set the average rank budget of IG-LoRA at 8 following~\citet{ma2023llmPruner}. 
More details and ablation of the implementation are shown in Appendix~\ref{app:methods_details} and Appendix~\ref{app:more_analysis}, respectively.

\paragraph{Baselines}
We compare the single-shot pruning performance\footnote{Without recovery fine-tuning.}of \ours{} against the following baselines: \textbf{Magnitude-SP} measures the importance criterion based on the magnitude of weights~\citep{han2015deep,jaiswal2023theemergence}.
This baseline employs uniform sparsity across blocks.
\textbf{Wanda-SP} is extended by the unstructured pruning method Wanda~\citep{wanda}, which modifies the target pruning units to structured weights.
We globally sort the pruning units across all blocks to identify redundant components, as this strategy tends to achieve better performance compared to adopting a local manner for individual blocks.
\textbf{FLAP}~\citep{flap} uses the stability of activations as an importance criterion, also applying a global sorting strategy.
Notably, for a fair comparison, all baselines are implemented to prune the intermediate dimensions of the FFN, which are the same as \ours{}.
Details are shown in Appendix~\ref{app:baseline_details}.

\subsection{Main Results}
\label{subsec:main_results}

\paragraph{Zero-shot Tasks}
We present a performance comparison of the LLaMA3 family in Tables~\ref{tab:zs_llama3_8b} and~\ref{tab:zs_llama3_70b}, as well as LLaMA2-13B in Table ~\ref{tab:zs_llama2_13b}.
In the single-shot pruning setting, \ours{} consistently demonstrates superior average performance compared to baselines across various models at both 20\% and 50\% sparsity.
Remarkably, \ours{} achieves a promising accuracy of 32.39 on MMLU with 50\% sparsity on LLaMA3-8B, while other baselines regress to chance-level accuracy (\textasciitilde25.0).
This result underscores the potential of \ours{} to perform well on more challenging tasks without retraining, even at high sparsity.
Furthermore, \ours{} is more favorable for larger models. 
At the 20\% and 50\% sparsity on LLaMA3-70B, \ours{} maintains 97.9\% and 82.2\% of the original performance on average, respectively.
We further evaluate \ours{} with recovery fine-tuning at 50\% sparsity for each model.
For comparison, we choose Wanda-SP, as it has the second-best average performance in single-shot pruning.
We fine-tune pruned models with our proposed IG-LoRA on 0.1B tokens from the FineWeb-Edu dataset.
We find that after recovery training, both pruning models are improved, especially on complex knowledge-sensitive tasks.
\ours{} still outperforms Wanda-SP in general, indicating the effectiveness of our proposed pruning and recovery approach.

\begin{table}[t]
\centering
\resizebox{1.0\linewidth}{!}{
\begin{tabular}{l|l|cc}
\toprule
Sparsity & Method & LLaMA3-8B & LLaMA3-70B \\
\midrule
 {0\%} & Dense & 6.82 & 5.26 \\
\midrule
    & \multicolumn{3}{l}{\textit{w/o recovery}} \\
 \multirow{2}{*}{20\%} &  Wanda-SP & 9.39 & \textbf{7.86} \\
&FLAP &  9.40 & 8.21\\
& \ours{}(ours) &  \textbf{8.97} & 8.02 \\
\midrule
    & \multicolumn{3}{l}{\textit{w/o recovery}} \\
\multirow{5}{*}{50\%} & Wanda-SP & $19.49$ & $13.53$ \\
&FLAP &  $21.06$ & $13.37$ \\
 & \ours{}(ours) &$\textbf{17.45}$ & $\textbf{13.02}$ \\
 \cmidrule(lr){2-4} 
    & \multicolumn{3}{l}{\textit{w/ recovery}} \\
& Wanda-SP &  $14.52$ & $11.75$  \\
& \ours{}(ours) &   $\textbf{12.55}$ & $\textbf{10.92}$ \\
\bottomrule
\end{tabular}
    }
\caption{Perplexity of pruning methods for LLaMA3-8B and LLaMA3-70B on WikiText2 validation set.}
\label{tab:ppl}
\end{table}
\paragraph{Language Modeling}
Table~\ref{tab:ppl} presents the perplexity on WikiText2. 
\ours{} consistently achieves better results than baselines, except for the 20\% sparsity on LLaMA3-8B, where it performs slightly worse than Wanda-SP.
Additionally, the benefits of \ours{} are more pronounced at higher sparsity.

\begin{table}[t]
\centering
\setlength{\tabcolsep}{2.5pt}
\small
\begin{tabular}{cccccc}
\toprule
\multirow{2}{*}{\textbf{Model}}  & 
\multirow{2}{*} {Parameters} & 
\multirow{2}{*} {Memory} & 
\multirow{2}{*} {MACs} & 
\multicolumn{2}{c}{Speed-up} \\
 & & & & CPU & GPU \\ 
\midrule
LLaMA3-8B & 8.03B  & 16.06GB & 3.64T & 1.0x & 1.0x\\
+ \ours{} & 5.21B  & 10.42GB & 2.19T & 2.3x & 1.6x\\
\midrule
LLaMA2-7B & 6.73B  & 12.61GB & 3.38T & 1.0x & 1.0x\\
+ \ours{} & 4.57B  & 8.62GB & 2.17T & 2.1x & 1.5x \\
\bottomrule
\end{tabular}
\caption{Comparison of parameter size, memory usage, MACs, and inference speed-up on CPU/GPU. The pruned models (+\ours{}) are under 50\% sparsity.}
\label{tab:analysis_efficient}
\end{table}

\subsection{Efficiency Evaluation}
We assess the inference efficiency of the pruned models.
The details of evaluation are shown in Appendix~\ref{app:benchmarks}.
The results of 50\% sparsity are shown in Table~\ref{tab:analysis_efficient}. 
Compared to the original dense models, \ours{} reduces the parameters, memory, and MACs by 40\% and achieves a speed-up over $1.5\times$ on CPU and GPU.
We also report the pruning and recovery time in Appendix~\ref{app:methods_details}.
In general, \ours{} significantly improves efficiency, indicating its effectiveness for practical deployments of LLM.

\begin{table}[t]
\centering
\small
\begin{tabular}{lccc}
\toprule
Setting & PPL↓ & HellaSwag & MMLU \\
\midrule
\multicolumn{4}{l}{\textit{coarse-grained importance ablation}} \\
(a) Uniform & 9.08 & 70.84 & 50.91 \\
(b) Euclidean & 9.11 & 70.11 & 50.19 \\
(c) Cosine & 8.98 & 72.52 & 55.79 \\
Angular (Ours) & \textbf{8.97} & \textbf{72.74} & \textbf{56.43} \\

\midrule
\multicolumn{4}{l}{\textit{fine-grained importance ablation}} \\
(d) Wanda & 9.03 & 71.93 & 55.33 \\
Eq~\eqref{eq:24} (Ours) & \textbf{8.97} & \textbf{72.74} & \textbf{56.43} \\

\bottomrule
\end{tabular}
\caption{Ablation of importance criterion of \ours{} on LLaMA3-8B under 20\% sparsity.
}
\label{tab:ablation_only_prune}
\end{table}

\subsection{Ablation Study}
\label{subsec:Ablation}

\paragraph{Importance Criterion}
We explore the effects of each component incorporated in the importance criterion of the proposed \ours{}.
Table~\ref{tab:ablation_only_prune} shows the ablation results under 20\% sparsity of LLaMA3-8B. 
We first investigate the coarse-grained importance of blocks by comparing variants including: (a) uniform sparsity for each block, (b) Euclidean distance, or (c) cosine similarity as the coarse-grained importance criterion to allocation sparsity budget across blocks.
As illustrated in Table~\ref{tab:ablation_only_prune}, applying uniform sparsity or using Euclidean distance results in a notable performance decrease, particularly for zero-shot tasks.
The angular distance (Eq.~\ref{eq:arccos-sim}) used in \ours{} achieves the best performance across tasks.
For fine-grained importance ablation, as shown in Table~\ref{tab:ablation_only_prune}, the criterion outlined in Eq.~\ref{eq:24} also demonstrates superior performance compared to the criterion utilized in Wanda.

\begin{figure}[t!]
\centering
\includegraphics[width=0.9\linewidth]{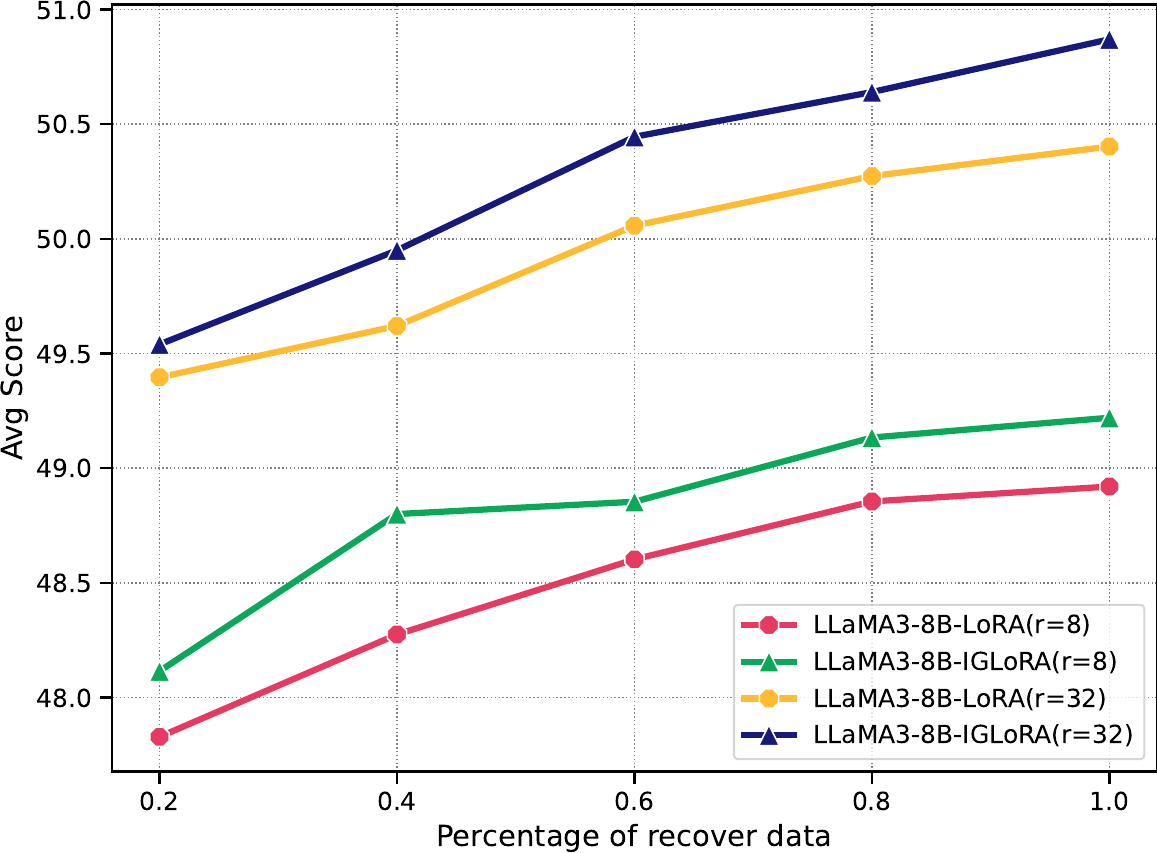}
\caption{
Results of different recovery fine-tuning methods at different data sizes.$r=$ 8/32 means the average rank budget configuration is set to 8 or 32.
}\label{fig:ours_vs_lora}
\end{figure}

\paragraph{Recovery Fine-tuning}
To assess the impact of our proposed IG-LoRA for recovery, we compare it with the original LoRA.
Figure~\ref{fig:ours_vs_lora} shows that IG-LoRA exhibits better performance than LoRA across various recovery data sizes and rank configurations. 
Furthermore, IG-LoRA achieves a performance comparable to LoRA trained on the full dataset while utilizing only 60\% of data, highlighting the efficiency of IG-LoRA.

\begin{figure}[t!]
\centering
\includegraphics[width=0.9\linewidth]{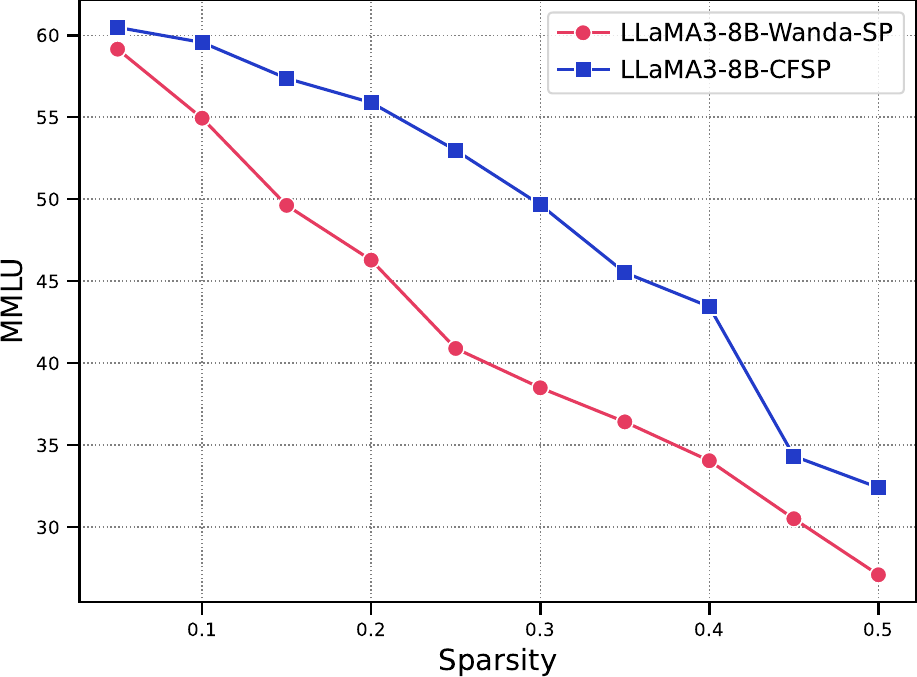}
\caption{
Performance comparison of MMLU task between \ours{} and Wanda-SP on LLaMA3-8B with various sparsity.
}\label{fig:various_sparsity_ratios}
\end{figure}

\subsection{Analysis}
\label{sec:analysis}
\paragraph{Performance with Various Sparsity}
Figure~\ref{fig:various_sparsity_ratios} presents the MMLU results of pruned models with sparsity from 5\% to 50\%. 
Under lower sparsity (10\%), Wanda-SP is comparable to \ours{}. As the sparsity increases, its performance decreases significantly, while \ours{} still maintains promising performance even at 50\% sparsity.

\begin{figure}[t!]
\centering
\includegraphics[width=1.0\linewidth]{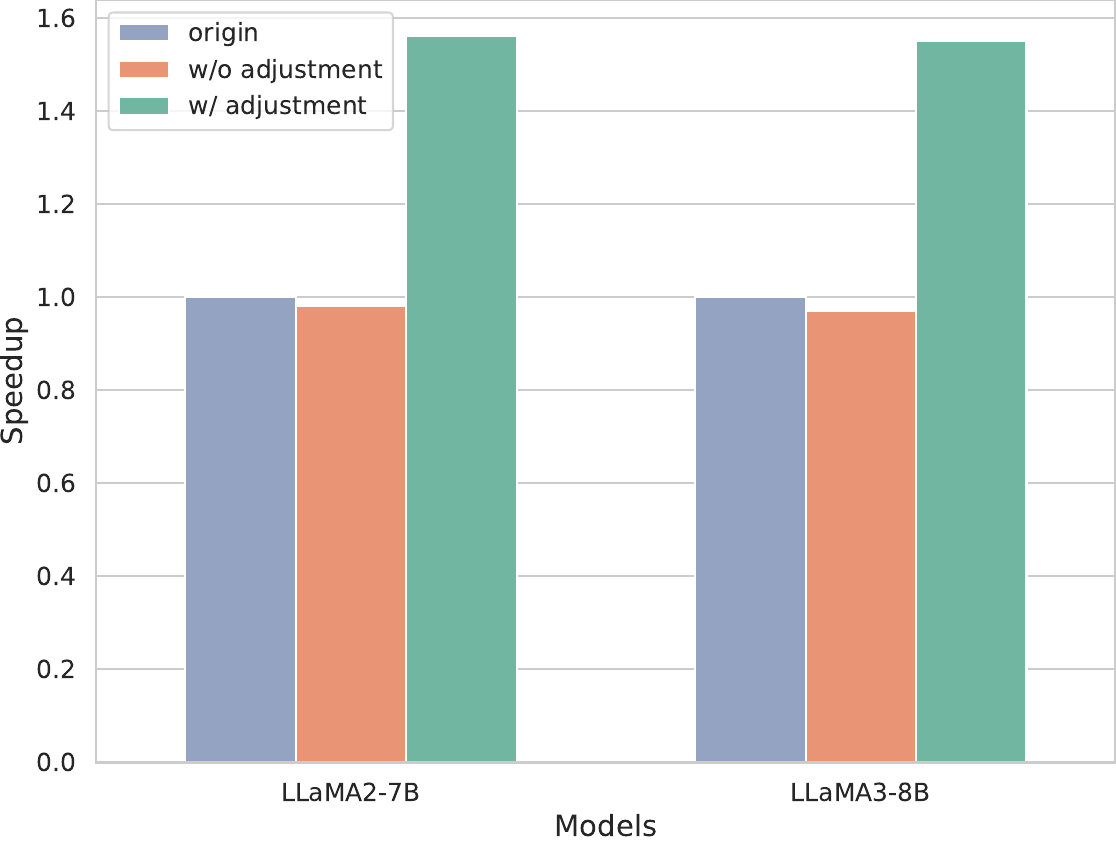}
\caption{
The effect of dimension adjustment. The speed-up is evaluated on a NVIDIA A800-80G.
}
\label{fig:speed}
\end{figure}

\paragraph{Impact of Dimension Adjustment}
Figure~\ref{fig:speed} compares the inference speed-up of whether to perform dimension adjustment during pruning. 
We observe that adjusting the intermediate dimension significantly accelerates models ($1.6\times$).
However, without the adjustment, the latency of pruned models is comparable to the original dense models.
Furthermore, the cost of adjustment is negligible and does not impact performance.
For instance, on LLaMA3-8B, the number of parameters increased only by 0.43\% after adjustment (from 5.21B to 5.23B).
The average zero-shot performance remains comparable to that without adjustment (41.67 \textit{vs} 41.61).

\paragraph{Visualization}

\begin{figure}[t!]
\centering
\includegraphics[width=0.9\linewidth]{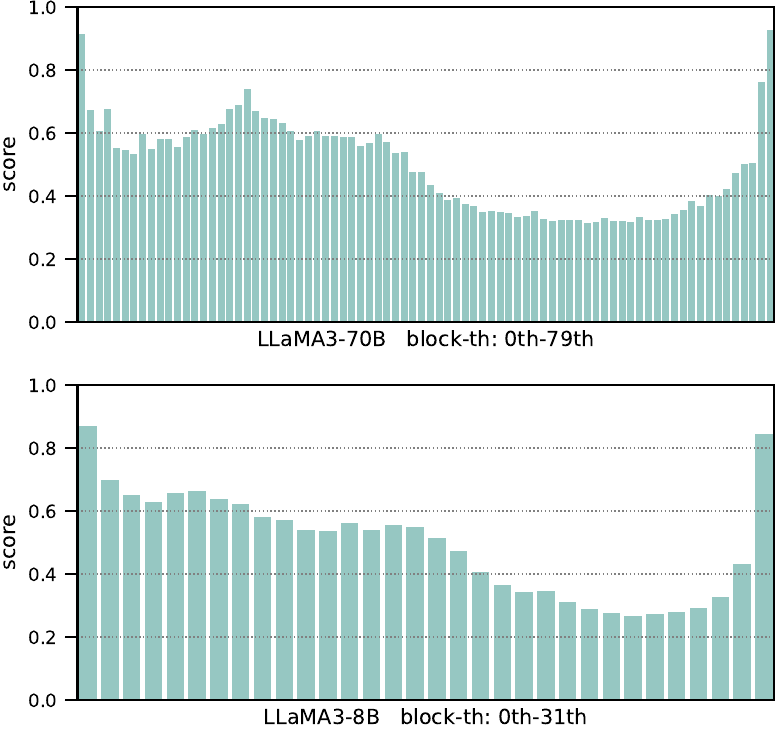}
\caption{
The visualizations of normalized coarse-grained importance scores of each block on LLaMA3-8B and LLaMA3-70B.
}\label{fig:distribution_of_coarse}
\end{figure}

We present a visualization of block importance scores on LLaMA3 in Figure~\ref{fig:distribution_of_coarse}. 
We find that the scores vary significantly across blocks: the first and last blocks exhibit the highest scores, whereas the intermediate blocks show lower scores. 
The varying importance explains why the uniform sparsity allocation is suboptimal and indicates that intermediate blocks exhibit greater redundancy, allowing for more aggressive pruning, while other blocks need to retain more weights.

\section{Related Work}
\paragraph{LLM Compression}
The enormous computations of LLMs has prompted efforts in improving their efficiency, including quantization~\citep{dettmers2022LLMint8, yao2022zeroquant, lin2024awq, frantar2022gptq, Xiao23smoothquant, dettmers2023qlora, dettmers2023spqr, shao2023omniquant, xu2024onebit}, distillation~\citep{wang2022distilled, jiang2023lion, wang2023scott, hsieh2023distilling, agarwal2023gkd, minillm} and KV cache compression~\citep{sheng2023flexgen, ge2023model, zhang2023h2o, liu2024kivi, kvquant, deepseekai2024deepseekv2}.
Pruning is another crucial method by eliminating redundant parameters.
Most of the previous pruning work follows the \textit{unstructured pruning}, which removes individual parameters according to their importance~\citep{sparsegpt, wanda,dettmers2023spqr, xu2024besa, zhang2024plugandplay}. 
However, this paradigm requires specialized hardware support to speed up.
In contrast, \textit{structured pruning} eliminates the structural group of weights, facilitating a more convenient deployment on general hardware~\citep{wang2020structured, xia2022structured}.
Some work proposes to remove redundant layers in LLMs~\citep{shortgpt, yang2024laco, UnreasonableIneffectiveness}, while dropping entire layers leads to a significant performance drop.
For pruning on more fine-grained units, some work formulates pruning as a constrained optimization problem by introducing learnable masks to search~\citep{ShearedLLaMA, bonsai, minitron2024, gunter2024apple}.
\citet{zhang2024structured} performs iteratively to prune the coupled weights until the desired sparsity is achieved.
\citet{ma2023llmPruner} and \citet{loraprune} use gradient information to guide pruning.
These methods incur substantial pruning overhead, particularly in the case of large-scale models.
\citet{flap} eliminates channels based on their activation fluctuations using only forward passes. 
In this work, we also aim to achieve efficient structured pruning using only forward passes.

\paragraph{Sparsity in Transformer}
Sparsity is a common trait in neural networks~\citep{zhu2019overparam, frankle2019lth, jaszczur2021sparse} and a lot of work explores sparsity in Transformer, such as attention~\citep{voita2019analyzing, michel2019are, hao2021self, zhu2021less} or FFN~\citep{ wang2020structured, zhang2022moefication, zuo2022moebert}.
The dynamic sparsity has also garnered attention~\citep{liu2023DejaVu, SmartTrim}, which adaptively selects a portion of the model based on input.
\citet{yin2024owl} find that non-uniform sparsity yields better results for LLM unstructured pruning, which is consistent with our observation in structured pruning.

\section{Conclusion}
In this work, we explore structured pruning for Large Language Models (LLMs).
We propose an efficient pruning framework named \ours{}, which leverages coarse to fine-grained activation information as an importance criterion to determine the redundant parts to prune.
For the coarse-grained importance, we measure the saliency of transformations of each block and use this criterion to assign the sparsity budget across blocks.
For weights within each block, we utilize a fine-grained criterion to remove redundant parts to obtain compact models.
We also introduce an efficient recovery fine-tuning method IG-LoRA that adaptively assigns additional trainable parameters based on the importance of blocks.
Extensive experimental results demonstrate that \ours{} outperforms existing methods across various models and sparsity levels, both in single-shot pruning and in recovery fine-tuning.
Meanwhile, even at high sparsity, our method can maintain promising performance on challenging tasks such as MMLU and FreebaseQA compared to the original dense models.

\section*{Limitations}
\ours{} is a fast and efficient structured pruning method for large language models (LLMs), while it also has some limitations. 
First, our experiments focus on the LLaMA family of models~\citep{llama1, llama2, llama3, llama3.1}, as they are among the most advanced open-source LLMs currently. 
We will extend our method to a broader range of models in the future.
Additionally, we do not prune attention heads, as this has been shown to cause significant performance degradation, especially for models that have grouped query attention (GQA)~\citep{gqa} like LLaMA3. Further research is needed to develop more effective pruning strategies, especially in the context of attention optimization techniques like GQA.

\section*{Acknowledgment}
The research in this article is supported by the National Science Foundation of China (U22B2059, 62276083), the Human-Machine Integrated Consultation System for Cardiovascular Diseases (2023A003).

\bibliography{custom}

\begin{thebibliography}{84}
\providecommand{\natexlab}[1]{#1}

\bibitem[{Agarwal et~al.(2023)Agarwal, Vieillard, Stanczyk, Ramos, Geist, and Bachem}]{agarwal2023gkd}
Rishabh Agarwal, Nino Vieillard, Piotr Stanczyk, Sabela Ramos, Matthieu Geist, and Olivier Bachem. 2023.
\newblock \href {https://arxiv.org/abs/2306.13649} {{GKD:} generalized knowledge distillation for auto-regressive sequence models}.
\newblock \emph{ArXiv preprint}.

\bibitem[{Ainslie et~al.(2023)Ainslie, Lee{-}Thorp, de~Jong, Zemlyanskiy, Lebr{\'{o}}n, and Sanghai}]{gqa}
Joshua Ainslie, James Lee{-}Thorp, Michiel de~Jong, Yury Zemlyanskiy, Federico Lebr{\'{o}}n, and Sumit Sanghai. 2023.
\newblock \href {https://doi.org/10.18653/v1/2023.emnlp-main.298} {{GQA:} training generalized multi-query transformer models from multi-head checkpoints}.
\newblock In \emph{Proceedings of the 2023 Conference on Empirical Methods in Natural Language Processing, {EMNLP} 2023, Singapore, December 6-10, 2023}.

\bibitem[{Allen{-}Zhu et~al.(2019)Allen{-}Zhu, Li, and Song}]{zhu2019overparam}
Zeyuan Allen{-}Zhu, Yuanzhi Li, and Zhao Song. 2019.
\newblock \href {http://proceedings.mlr.press/v97/allen-zhu19a.html} {A convergence theory for deep learning via over-parameterization}.
\newblock In \emph{Proc. of ICML}, Proceedings of Machine Learning Research.

\bibitem[{An et~al.(2024)An, Zhao, Yu, Tang, and Wang}]{flap}
Yongqi An, Xu~Zhao, Tao Yu, Ming Tang, and Jinqiao Wang. 2024.
\newblock \href {https://doi.org/10.1609/aaai.v38i10.28960} {Fluctuation-based adaptive structured pruning for large language models}.
\newblock In \emph{Thirty-Eighth {AAAI} Conference on Artificial Intelligence, {AAAI} 2024, Thirty-Sixth Conference on Innovative Applications of Artificial Intelligence, {IAAI} 2024, Fourteenth Symposium on Educational Advances in Artificial Intelligence, {EAAI} 2014, February 20-27, 2024, Vancouver, Canada}.

\bibitem[{Ashkboos et~al.(2024)Ashkboos, Croci, Nascimento, Hoefler, and Hensman}]{ashkboos2024slicegpt}
Saleh Ashkboos, Maximilian~L. Croci, Marcelo Gennari~Do Nascimento, Torsten Hoefler, and James Hensman. 2024.
\newblock \href {https://openreview.net/forum?id=vXxardq6db} {Slicegpt: Compress large language models by deleting rows and columns}.
\newblock In \emph{Proc. of ICLR}.

\bibitem[{Bisk et~al.(2020)Bisk, Zellers, LeBras, Gao, and Choi}]{PIQA}
Yonatan Bisk, Rowan Zellers, Ronan LeBras, Jianfeng Gao, and Yejin Choi. 2020.
\newblock \href {https://aaai.org/ojs/index.php/AAAI/article/view/6239} {{PIQA:} reasoning about physical commonsense in natural language}.
\newblock In \emph{The Thirty-Fourth {AAAI} Conference on Artificial Intelligence, {AAAI} 2020, The Thirty-Second Innovative Applications of Artificial Intelligence Conference, {IAAI} 2020, The Tenth {AAAI} Symposium on Educational Advances in Artificial Intelligence, {EAAI} 2020, New York, NY, USA, February 7-12, 2020}.

\bibitem[{Brown et~al.(2020)Brown, Mann, Ryder, Subbiah, Kaplan, Dhariwal, Neelakantan, Shyam, Sastry, Askell, Agarwal, Herbert{-}Voss, Krueger, Henighan, Child, Ramesh, Ziegler, Wu, Winter, Hesse, Chen, Sigler, Litwin, Gray, Chess, Clark, Berner, McCandlish, Radford, Sutskever, and Amodei}]{gpt3}
Tom~B. Brown, Benjamin Mann, Nick Ryder, Melanie Subbiah, Jared Kaplan, Prafulla Dhariwal, Arvind Neelakantan, Pranav Shyam, Girish Sastry, Amanda Askell, Sandhini Agarwal, Ariel Herbert{-}Voss, Gretchen Krueger, Tom Henighan, Rewon Child, Aditya Ramesh, Daniel~M. Ziegler, Jeffrey Wu, Clemens Winter, Christopher Hesse, Mark Chen, Eric Sigler, Mateusz Litwin, Scott Gray, Benjamin Chess, Jack Clark, Christopher Berner, Sam McCandlish, Alec Radford, Ilya Sutskever, and Dario Amodei. 2020.
\newblock \href {https://proceedings.neurips.cc/paper/2020/hash/1457c0d6bfcb4967418bfb8ac142f64a-Abstract.html} {Language models are few-shot learners}.
\newblock In \emph{Advances in Neural Information Processing Systems 33: Annual Conference on Neural Information Processing Systems 2020, NeurIPS 2020, December 6-12, 2020, virtual}.

\bibitem[{Clark et~al.(2018)Clark, Cowhey, Etzioni, Khot, Sabharwal, Schoenick, and Tafjord}]{arc}
Peter Clark, Isaac Cowhey, Oren Etzioni, Tushar Khot, Ashish Sabharwal, Carissa Schoenick, and Oyvind Tafjord. 2018.
\newblock \href {https://arxiv.org/abs/1803.05457} {Think you have solved question answering? try arc, the {AI2} reasoning challenge}.
\newblock \emph{ArXiv preprint}.

\bibitem[{Computer(2023)}]{together2023redpajama}
Together Computer. 2023.
\newblock \href {https://github.com/togethercomputer/RedPajama-Data} {Redpajama: an open dataset for training large language models}.

\bibitem[{DeepSeek-AI et~al.(2024)DeepSeek-AI, Liu, Feng, Wang, Wang, Liu, Zhao, Dengr, Ruan, Dai, Guo, Yang, Chen, Ji, Li, Lin, Luo, Hao, Chen, Li, Zhang, Xu, Yang, Zhang, Ding, Xin, Gao, Li, Qu, Cai, Liang, Guo, Ni, Li, Chen, Yuan, Qiu, Song, Dong, Gao, Guan, Wang, Zhang, Xu, Xia, Zhao, Zhang, Li, Wang, Zhang, Zhang, Tang, Li, Tian, Huang, Wang, Zhang, Zhu, Chen, Du, Chen, Jin, Ge, Pan, Xu, Chen, Li, Lu, Zhou, Chen, Wu, Ye, Ma, Wang, Zhou, Yu, Zhou, Zheng, Wang, Pei, Yuan, Sun, Xiao, Zeng, An, Liu, Liang, Gao, Zhang, Li, Jin, Wang, Bi, Liu, Wang, Shen, Chen, Chen, Nie, Sun, Wang, Liu, Xie, Yu, Song, Zhou, Yang, Lu, Su, Wu, Li, Wei, Zhu, Xu, Huang, Li, Zhao, Sun, Li, Wang, Zheng, Zhang, Xiong, Zhao, He, Tang, Piao, Dong, Tan, Liu, Wang, Guo, Zhu, Wang, Zou, Zha, Ma, Yan, You, Liu, Ren, Ren, Sha, Fu, Huang, Zhang, Xie, Hao, Shao, Wen, Xu, Zhang, Li, Wang, Gu, Li, and Xie}]{deepseekai2024deepseekv2}
DeepSeek-AI, Aixin Liu, Bei Feng, Bin Wang, Bingxuan Wang, Bo~Liu, Chenggang Zhao, Chengqi Dengr, Chong Ruan, Damai Dai, Daya Guo, Dejian Yang, Deli Chen, Dongjie Ji, Erhang Li, Fangyun Lin, Fuli Luo, Guangbo Hao, Guanting Chen, Guowei Li, H.~Zhang, Hanwei Xu, Hao Yang, Haowei Zhang, Honghui Ding, Huajian Xin, Huazuo Gao, Hui Li, Hui Qu, J.~L. Cai, Jian Liang, Jianzhong Guo, Jiaqi Ni, Jiashi Li, Jin Chen, Jingyang Yuan, Junjie Qiu, Junxiao Song, Kai Dong, Kaige Gao, Kang Guan, Lean Wang, Lecong Zhang, Lei Xu, Leyi Xia, Liang Zhao, Liyue Zhang, Meng Li, Miaojun Wang, Mingchuan Zhang, Minghua Zhang, Minghui Tang, Mingming Li, Ning Tian, Panpan Huang, Peiyi Wang, Peng Zhang, Qihao Zhu, Qinyu Chen, Qiushi Du, R.~J. Chen, R.~L. Jin, Ruiqi Ge, Ruizhe Pan, Runxin Xu, Ruyi Chen, S.~S. Li, Shanghao Lu, Shangyan Zhou, Shanhuang Chen, Shaoqing Wu, Shengfeng Ye, Shirong Ma, Shiyu Wang, Shuang Zhou, Shuiping Yu, Shunfeng Zhou, Size Zheng, T.~Wang, Tian Pei, Tian Yuan, Tianyu Sun, W.~L. Xiao, Wangding Zeng, Wei An, Wen
  Liu, Wenfeng Liang, Wenjun Gao, Wentao Zhang, X.~Q. Li, Xiangyue Jin, Xianzu Wang, Xiao Bi, Xiaodong Liu, Xiaohan Wang, Xiaojin Shen, Xiaokang Chen, Xiaosha Chen, Xiaotao Nie, Xiaowen Sun, Xiaoxiang Wang, Xin Liu, Xin Xie, Xingkai Yu, Xinnan Song, Xinyi Zhou, Xinyu Yang, Xuan Lu, Xuecheng Su, Y.~Wu, Y.~K. Li, Y.~X. Wei, Y.~X. Zhu, Yanhong Xu, Yanping Huang, Yao Li, Yao Zhao, Yaofeng Sun, Yaohui Li, Yaohui Wang, Yi~Zheng, Yichao Zhang, Yiliang Xiong, Yilong Zhao, Ying He, Ying Tang, Yishi Piao, Yixin Dong, Yixuan Tan, Yiyuan Liu, Yongji Wang, Yongqiang Guo, Yuchen Zhu, Yuduan Wang, Yuheng Zou, Yukun Zha, Yunxian Ma, Yuting Yan, Yuxiang You, Yuxuan Liu, Z.~Z. Ren, Zehui Ren, Zhangli Sha, Zhe Fu, Zhen Huang, Zhen Zhang, Zhenda Xie, Zhewen Hao, Zhihong Shao, Zhiniu Wen, Zhipeng Xu, Zhongyu Zhang, Zhuoshu Li, Zihan Wang, Zihui Gu, Zilin Li, and Ziwei Xie. 2024.
\newblock \href {https://arxiv.org/abs/2405.04434} {Deepseek-v2: A strong, economical, and efficient mixture-of-experts language model}.
\newblock \emph{Preprint}, arXiv:2405.04434.

\bibitem[{Dery et~al.(2024)Dery, Kolawole, Kagey, Smith, Neubig, and Talwalkar}]{bonsai}
Lucio~M. Dery, Steven Kolawole, Jean{-}Fran{\c{c}}ois Kagey, Virginia Smith, Graham Neubig, and Ameet Talwalkar. 2024.
\newblock \href {https://arxiv.org/abs/2402.05406} {Everybody prune now: Structured pruning of llms with only forward passes}.
\newblock \emph{ArXiv preprint}.

\bibitem[{Dettmers et~al.(2022)Dettmers, Lewis, Belkada, and Zettlemoyer}]{dettmers2022LLMint8}
Tim Dettmers, Mike Lewis, Younes Belkada, and Luke Zettlemoyer. 2022.
\newblock \href {https://arxiv.org/abs/2208.07339} {Llm.int8(): 8-bit matrix multiplication for transformers at scale}.
\newblock \emph{ArXiv preprint}.

\bibitem[{Dettmers et~al.(2023{\natexlab{a}})Dettmers, Pagnoni, Holtzman, and Zettlemoyer}]{dettmers2023qlora}
Tim Dettmers, Artidoro Pagnoni, Ari Holtzman, and Luke Zettlemoyer. 2023{\natexlab{a}}.
\newblock \href {http://papers.nips.cc/paper\_files/paper/2023/hash/1feb87871436031bdc0f2beaa62a049b-Abstract-Conference.html} {Qlora: Efficient finetuning of quantized llms}.
\newblock In \emph{Advances in Neural Information Processing Systems 36: Annual Conference on Neural Information Processing Systems 2023, NeurIPS 2023, New Orleans, LA, USA, December 10 - 16, 2023}.

\bibitem[{Dettmers et~al.(2023{\natexlab{b}})Dettmers, Svirschevski, Egiazarian, Kuznedelev, Frantar, Ashkboos, Borzunov, Hoefler, and Alistarh}]{dettmers2023spqr}
Tim Dettmers, Ruslan Svirschevski, Vage Egiazarian, Denis Kuznedelev, Elias Frantar, Saleh Ashkboos, Alexander Borzunov, Torsten Hoefler, and Dan Alistarh. 2023{\natexlab{b}}.
\newblock \href {https://arxiv.org/abs/2306.03078} {Spqr: {A} sparse-quantized representation for near-lossless {LLM} weight compression}.
\newblock \emph{ArXiv preprint}.

\bibitem[{Dubey et~al.(2024)Dubey, Jauhri, Pandey, Kadian, Al{-}Dahle, Letman, Mathur, Schelten, Yang, Fan, Goyal, Hartshorn, Yang, Mitra, Sravankumar, Korenev, Hinsvark, Rao, Zhang, Rodriguez, Gregerson, Spataru, Rozi{\`{e}}re, Biron, Tang, Chern, Caucheteux, Nayak, Bi, Marra, McConnell, Keller, Touret, Wu, Wong, Ferrer, Nikolaidis, Allonsius, Song, Pintz, Livshits, Esiobu, Choudhary, Mahajan, Garcia{-}Olano, Perino, Hupkes, Lakomkin, AlBadawy, Lobanova, Dinan, Smith, Radenovic, Zhang, Synnaeve, Lee, Anderson, Nail, Mialon, Pang, Cucurell, Nguyen, Korevaar, Xu, Touvron, Zarov, Ibarra, Kloumann, Misra, Evtimov, Copet, Lee, Geffert, Vranes, Park, Mahadeokar, Shah, van~der Linde, Billock, Hong, Lee, Fu, Chi, Huang, Liu, Wang, Yu, Bitton, Spisak, Park, Rocca, Johnstun, Saxe, Jia, Alwala, Upasani, Plawiak, Li, Heafield, Stone, and et~al.}]{llama3.1}
Abhimanyu Dubey, Abhinav Jauhri, Abhinav Pandey, Abhishek Kadian, Ahmad Al{-}Dahle, Aiesha Letman, Akhil Mathur, Alan Schelten, Amy Yang, Angela Fan, Anirudh Goyal, Anthony Hartshorn, Aobo Yang, Archi Mitra, Archie Sravankumar, Artem Korenev, Arthur Hinsvark, Arun Rao, Aston Zhang, Aur{\'{e}}lien Rodriguez, Austen Gregerson, Ava Spataru, Baptiste Rozi{\`{e}}re, Bethany Biron, Binh Tang, Bobbie Chern, Charlotte Caucheteux, Chaya Nayak, Chloe Bi, Chris Marra, Chris McConnell, Christian Keller, Christophe Touret, Chunyang Wu, Corinne Wong, Cristian~Canton Ferrer, Cyrus Nikolaidis, Damien Allonsius, Daniel Song, Danielle Pintz, Danny Livshits, David Esiobu, Dhruv Choudhary, Dhruv Mahajan, Diego Garcia{-}Olano, Diego Perino, Dieuwke Hupkes, Egor Lakomkin, Ehab AlBadawy, Elina Lobanova, Emily Dinan, Eric~Michael Smith, Filip Radenovic, Frank Zhang, Gabriel Synnaeve, Gabrielle Lee, Georgia~Lewis Anderson, Graeme Nail, Gr{\'{e}}goire Mialon, Guan Pang, Guillem Cucurell, Hailey Nguyen, Hannah Korevaar, Hu~Xu, Hugo
  Touvron, Iliyan Zarov, Imanol~Arrieta Ibarra, Isabel~M. Kloumann, Ishan Misra, Ivan Evtimov, Jade Copet, Jaewon Lee, Jan Geffert, Jana Vranes, Jason Park, Jay Mahadeokar, Jeet Shah, Jelmer van~der Linde, Jennifer Billock, Jenny Hong, Jenya Lee, Jeremy Fu, Jianfeng Chi, Jianyu Huang, Jiawen Liu, Jie Wang, Jiecao Yu, Joanna Bitton, Joe Spisak, Jongsoo Park, Joseph Rocca, Joshua Johnstun, Joshua Saxe, Junteng Jia, Kalyan~Vasuden Alwala, Kartikeya Upasani, Kate Plawiak, Ke~Li, Kenneth Heafield, Kevin Stone, and et~al. 2024.
\newblock \href {https://arxiv.org/abs/2407.21783} {The llama 3 herd of models}.
\newblock \emph{ArXiv preprint}.

\bibitem[{Fei et~al.(2024)Fei, Shao, Li, Zeng, Yan, Qiu, and Lin}]{fei2024query}
Zhaoye Fei, Yunfan Shao, Linyang Li, Zhiyuan Zeng, Hang Yan, Xipeng Qiu, and Dahua Lin. 2024.
\newblock \href {https://arxiv.org/abs/2401.14624} {Query of cc: Unearthing large scale domain-specific knowledge from public corpora}.
\newblock \emph{ArXiv preprint}.

\bibitem[{Frankle and Carbin(2019)}]{frankle2019lth}
Jonathan Frankle and Michael Carbin. 2019.
\newblock \href {https://openreview.net/forum?id=rJl-b3RcF7} {The lottery ticket hypothesis: Finding sparse, trainable neural networks}.
\newblock In \emph{Proc. of ICLR}.

\bibitem[{Frantar and Alistarh(2023)}]{sparsegpt}
Elias Frantar and Dan Alistarh. 2023.
\newblock \href {https://proceedings.mlr.press/v202/frantar23a.html} {Sparsegpt: Massive language models can be accurately pruned in one-shot}.
\newblock In \emph{International Conference on Machine Learning, {ICML} 2023, 23-29 July 2023, Honolulu, Hawaii, {USA}}, Proceedings of Machine Learning Research.

\bibitem[{Frantar et~al.(2022)Frantar, Ashkboos, Hoefler, and Alistarh}]{frantar2022gptq}
Elias Frantar, Saleh Ashkboos, Torsten Hoefler, and Dan Alistarh. 2022.
\newblock \href {https://arxiv.org/abs/2210.17323} {{GPTQ:} accurate post-training quantization for generative pre-trained transformers}.
\newblock \emph{ArXiv preprint}.

\bibitem[{Gao et~al.(2023)Gao, Tow, Abbasi, Biderman, Black, DiPofi, Foster, Golding, Hsu, Le~Noac'h, Li, McDonell, Muennighoff, Ociepa, Phang, Reynolds, Schoelkopf, Skowron, Sutawika, Tang, Thite, Wang, Wang, and Zou}]{eval-harness}
Leo Gao, Jonathan Tow, Baber Abbasi, Stella Biderman, Sid Black, Anthony DiPofi, Charles Foster, Laurence Golding, Jeffrey Hsu, Alain Le~Noac'h, Haonan Li, Kyle McDonell, Niklas Muennighoff, Chris Ociepa, Jason Phang, Laria Reynolds, Hailey Schoelkopf, Aviya Skowron, Lintang Sutawika, Eric Tang, Anish Thite, Ben Wang, Kevin Wang, and Andy Zou. 2023.
\newblock \href {https://zenodo.org/records/10256836} {A framework for few-shot language model evaluation}.

\bibitem[{Ge et~al.(2023)Ge, Zhang, Liu, Zhang, Han, and Gao}]{ge2023model}
Suyu Ge, Yunan Zhang, Liyuan Liu, Minjia Zhang, Jiawei Han, and Jianfeng Gao. 2023.
\newblock \href {https://arxiv.org/abs/2310.01801} {Model tells you what to discard: Adaptive {KV} cache compression for llms}.
\newblock \emph{ArXiv preprint}.

\bibitem[{{Gemini Team} et~al.(2023){Gemini Team}, Anil, Borgeaud, Wu, Alayrac, Yu, Soricut, Schalkwyk, Dai, Hauth et~al.}]{team2023gemini}
{Gemini Team}, Rohan Anil, Sebastian Borgeaud, Yonghui Wu, Jean-Baptiste Alayrac, Jiahui Yu, Radu Soricut, Johan Schalkwyk, Andrew~M Dai, Anja Hauth, et~al. 2023.
\newblock \href {https://arxiv.org/abs/2312.11805} {Gemini: a family of highly capable multimodal models}.
\newblock \emph{ArXiv preprint}.

\bibitem[{Gromov et~al.(2024{\natexlab{a}})Gromov, Tirumala, Shapourian, Glorioso, and Roberts}]{gromov2024the}
Andrey Gromov, Kushal Tirumala, Hassan Shapourian, Paolo Glorioso, and Daniel~A. Roberts. 2024{\natexlab{a}}.
\newblock \href {https://arxiv.org/abs/2403.17887} {The unreasonable ineffectiveness of the deeper layers}.
\newblock \emph{ArXiv preprint}.

\bibitem[{Gromov et~al.(2024{\natexlab{b}})Gromov, Tirumala, Shapourian, Glorioso, and Roberts}]{UnreasonableIneffectiveness}
Andrey Gromov, Kushal Tirumala, Hassan Shapourian, Paolo Glorioso, and Daniel~A. Roberts. 2024{\natexlab{b}}.
\newblock \href {https://arxiv.org/abs/2403.17887} {The unreasonable ineffectiveness of the deeper layers}.
\newblock \emph{ArXiv preprint}.

\bibitem[{Gu et~al.(2023)Gu, Dong, Wei, and Huang}]{minillm}
Yuxian Gu, Li~Dong, Furu Wei, and Minlie Huang. 2023.
\newblock \href {https://arxiv.org/abs/2306.08543} {Knowledge distillation of large language models}.
\newblock \emph{ArXiv preprint}.

\bibitem[{Gunter et~al.(2024)Gunter, Wang, Wang, Pang, Narayanan, Zhang, Zhang, Chen, Chiu, Qiu, Gopinath, Yap, Yin, Nan, Weers, Yin, Huang, Wang, Lu, Peebles, Ye, Lee, Du, Chen, Keunebroek, Wiseman, Evans, Lei, Rathod, Kong, Du, Li, Wang, Gao, Ahmed, Xu, Lu, Rashid, Jose, Doane, Bencomo, Vanderby, Hansen, Jain, Anupama, Kamal, Wu, Brum, Maalouf, Erdenebileg, Dulhanty, Moritz, Kang, Jimenez, Ladd, Shi, Bai, Chu, Hohman, Kotek, Coleman, Li, Bigham, Cao, Lai, Cheung, Shan, Zhou, Li, Qin, Singh, Vega, Zou, Heckman, Gardiner, Bowler, Cordell, Cao, Hay, Shahdadpuri, Godwin, Dighe, Rachapudi, Tantawi, Frigg, Davarnia, Shah, Guha, Sirovica, Ma, Ma, Wang, Kim, Jayaram, Shankar, Paidi, Kumar, Wang, Zheng, and Cheng}]{gunter2024apple}
Tom Gunter, Zirui Wang, Chong Wang, Ruoming Pang, Andy Narayanan, Aonan Zhang, Bowen Zhang, Chen Chen, Chung{-}Cheng Chiu, David Qiu, Deepak Gopinath, Dian~Ang Yap, Dong Yin, Feng Nan, Floris Weers, Guoli Yin, Haoshuo Huang, Jianyu Wang, Jiarui Lu, John Peebles, Ke~Ye, Mark Lee, Nan Du, Qibin Chen, Quentin Keunebroek, Sam Wiseman, Syd Evans, Tao Lei, Vivek Rathod, Xiang Kong, Xianzhi Du, Yanghao Li, Yongqiang Wang, Yuan Gao, Zaid Ahmed, Zhaoyang Xu, Zhiyun Lu, Al~Rashid, Albin~Madappally Jose, Alec Doane, Alfredo Bencomo, Allison Vanderby, Andrew Hansen, Ankur Jain, Anupama~Mann Anupama, Areeba Kamal, Bugu Wu, Carolina Brum, Charlie Maalouf, Chinguun Erdenebileg, Chris Dulhanty, Dominik Moritz, Doug Kang, Eduardo Jimenez, Evan Ladd, Fangping Shi, Felix Bai, Frank Chu, Fred Hohman, Hadas Kotek, Hannah~Gillis Coleman, Jane Li, Jeffrey~P. Bigham, Jeffery Cao, Jeff Lai, Jessica Cheung, Jiulong Shan, Joe Zhou, John Li, Jun Qin, Karanjeet Singh, Karla Vega, Kelvin Zou, Laura Heckman, Lauren Gardiner, Margit Bowler,
  Maria Cordell, Meng Cao, Nicole Hay, Nilesh Shahdadpuri, Otto Godwin, Pranay Dighe, Pushyami Rachapudi, Ramsey Tantawi, Roman Frigg, Sam Davarnia, Sanskruti Shah, Saptarshi Guha, Sasha Sirovica, Shen Ma, Shuang Ma, Simon Wang, Sulgi Kim, Suma Jayaram, Vaishaal Shankar, Varsha Paidi, Vivek Kumar, Xin Wang, Xin Zheng, and Walker Cheng. 2024.
\newblock \href {https://arxiv.org/abs/2407.21075} {Apple intelligence foundation language models}.
\newblock \emph{ArXiv preprint}.

\bibitem[{Han et~al.(2016)Han, Mao, and Dally}]{han2015deep}
Song Han, Huizi Mao, and William~J. Dally. 2016.
\newblock \href {http://arxiv.org/abs/1510.00149} {Deep compression: Compressing deep neural network with pruning, trained quantization and huffman coding}.
\newblock In \emph{Proc. of ICLR}.

\bibitem[{Hao et~al.(2021)Hao, Dong, Wei, and Xu}]{hao2021self}
Yaru Hao, Li~Dong, Furu Wei, and Ke~Xu. 2021.
\newblock \href {https://ojs.aaai.org/index.php/AAAI/article/view/17533} {Self-attention attribution: Interpreting information interactions inside transformer}.
\newblock In \emph{Thirty-Fifth {AAAI} Conference on Artificial Intelligence, {AAAI} 2021, Thirty-Third Conference on Innovative Applications of Artificial Intelligence, {IAAI} 2021, The Eleventh Symposium on Educational Advances in Artificial Intelligence, {EAAI} 2021, Virtual Event, February 2-9, 2021}.

\bibitem[{Hendrycks et~al.(2021)Hendrycks, Burns, Basart, Zou, Mazeika, Song, and Steinhardt}]{mmlu}
Dan Hendrycks, Collin Burns, Steven Basart, Andy Zou, Mantas Mazeika, Dawn Song, and Jacob Steinhardt. 2021.
\newblock \href {https://openreview.net/forum?id=d7KBjmI3GmQ} {Measuring massive multitask language understanding}.
\newblock In \emph{Proc. of ICLR}.

\bibitem[{Hooper et~al.(2024)Hooper, Kim, Mohammadzadeh, Mahoney, Shao, Keutzer, and Gholami}]{kvquant}
Coleman Hooper, Sehoon Kim, Hiva Mohammadzadeh, Michael~W. Mahoney, Yakun~Sophia Shao, Kurt Keutzer, and Amir Gholami. 2024.
\newblock \href {https://arxiv.org/abs/2401.18079} {Kvquant: Towards 10 million context length {LLM} inference with {KV} cache quantization}.
\newblock \emph{ArXiv preprint}.

\bibitem[{Hsieh et~al.(2023)Hsieh, Li, Yeh, Nakhost, Fujii, Ratner, Krishna, Lee, and Pfister}]{hsieh2023distilling}
Cheng{-}Yu Hsieh, Chun{-}Liang Li, Chih{-}Kuan Yeh, Hootan Nakhost, Yasuhisa Fujii, Alex Ratner, Ranjay Krishna, Chen{-}Yu Lee, and Tomas Pfister. 2023.
\newblock \href {https://doi.org/10.18653/v1/2023.findings-acl.507} {Distilling step-by-step! outperforming larger language models with less training data and smaller model sizes}.
\newblock In \emph{Findings of the Association for Computational Linguistics: {ACL} 2023, Toronto, Canada, July 9-14, 2023}.

\bibitem[{Hu et~al.(2022)Hu, Shen, Wallis, Allen{-}Zhu, Li, Wang, Wang, and Chen}]{hu2022lora}
Edward~J. Hu, Yelong Shen, Phillip Wallis, Zeyuan Allen{-}Zhu, Yuanzhi Li, Shean Wang, Lu~Wang, and Weizhu Chen. 2022.
\newblock \href {https://openreview.net/forum?id=nZeVKeeFYf9} {Lora: Low-rank adaptation of large language models}.
\newblock In \emph{Proc. of ICLR}.

\bibitem[{Jaiswal et~al.(2024)Jaiswal, Gan, Du, Zhang, Wang, and Yang}]{llmkick}
Ajay Jaiswal, Zhe Gan, Xianzhi Du, Bowen Zhang, Zhangyang Wang, and Yinfei Yang. 2024.
\newblock Compressing llms: The truth is rarely pure and never simple.
\newblock In \emph{The Twelfth International Conference on Learning Representations}.

\bibitem[{Jaiswal et~al.(2023)Jaiswal, Liu, Chen, and Wang}]{jaiswal2023theemergence}
Ajay Jaiswal, Shiwei Liu, Tianlong Chen, and Zhangyang Wang. 2023.
\newblock \href {http://papers.nips.cc/paper\_files/paper/2023/hash/7a69ab48efcbb0153e72d458fb091969-Abstract-Conference.html} {The emergence of essential sparsity in large pre-trained models: The weights that matter}.
\newblock In \emph{Advances in Neural Information Processing Systems 36: Annual Conference on Neural Information Processing Systems 2023, NeurIPS 2023, New Orleans, LA, USA, December 10 - 16, 2023}.

\bibitem[{Jaszczur et~al.(2021)Jaszczur, Chowdhery, Mohiuddin, Kaiser, Gajewski, Michalewski, and Kanerva}]{jaszczur2021sparse}
Sebastian Jaszczur, Aakanksha Chowdhery, Afroz Mohiuddin, Lukasz Kaiser, Wojciech Gajewski, Henryk Michalewski, and Jonni Kanerva. 2021.
\newblock \href {https://proceedings.neurips.cc/paper/2021/hash/51f15efdd170e6043fa02a74882f0470-Abstract.html} {Sparse is enough in scaling transformers}.
\newblock In \emph{Advances in Neural Information Processing Systems 34: Annual Conference on Neural Information Processing Systems 2021, NeurIPS 2021, December 6-14, 2021, virtual}.

\bibitem[{Jiang et~al.(2019)Jiang, Wu, and Jiang}]{freebaseqa}
Kelvin Jiang, Dekun Wu, and Hui Jiang. 2019.
\newblock \href {https://aclanthology.org/N19-1028} {{F}reebase{QA}: A new factoid {QA} data set matching trivia-style question-answer pairs with {F}reebase}.
\newblock In \emph{Proc. of NAACL-HLT}.

\bibitem[{Jiang et~al.(2023)Jiang, Chan, Chen, and Wang}]{jiang2023lion}
Yuxin Jiang, Chunkit Chan, Mingyang Chen, and Wei Wang. 2023.
\newblock \href {https://doi.org/10.18653/v1/2023.emnlp-main.189} {Lion: Adversarial distillation of proprietary large language models}.
\newblock In \emph{Proceedings of the 2023 Conference on Empirical Methods in Natural Language Processing, {EMNLP} 2023, Singapore, December 6-10, 2023}.

\bibitem[{Lin et~al.(2024)Lin, Tang, Tang, Yang, Chen, Wang, Xiao, Dang, Gan, and Han}]{lin2024awq}
Ji~Lin, Jiaming Tang, Haotian Tang, Shang Yang, Wei{-}Ming Chen, Wei{-}Chen Wang, Guangxuan Xiao, Xingyu Dang, Chuang Gan, and Song Han. 2024.
\newblock \href {https://proceedings.mlsys.org/paper\_files/paper/2024/hash/42a452cbafa9dd64e9ba4aa95cc1ef21-Abstract-Conference.html} {{AWQ:} activation-aware weight quantization for on-device {LLM} compression and acceleration}.
\newblock In \emph{Proceedings of the Seventh Annual Conference on Machine Learning and Systems, MLSys 2024, Santa Clara, CA, USA, May 13-16, 2024}.

\bibitem[{Liu et~al.(2023)Liu, Wang, Dao, Zhou, Yuan, Song, Shrivastava, Zhang, Tian, Re, and Chen}]{liu2023DejaVu}
Zichang Liu, Jue Wang, Tri Dao, Tianyi Zhou, Binhang Yuan, Zhao Song, Anshumali Shrivastava, Ce~Zhang, Yuandong Tian, Christopher Re, and Beidi Chen. 2023.
\newblock \href {https://proceedings.mlr.press/v202/liu23am.html} {Deja vu: Contextual sparsity for efficient llms at inference time}.
\newblock In \emph{International Conference on Machine Learning, {ICML} 2023, 23-29 July 2023, Honolulu, Hawaii, {USA}}, Proceedings of Machine Learning Research.

\bibitem[{Liu et~al.(2024)Liu, Yuan, Jin, Zhong, Xu, Braverman, Chen, and Hu}]{liu2024kivi}
Zirui Liu, Jiayi Yuan, Hongye Jin, Shaochen Zhong, Zhaozhuo Xu, Vladimir Braverman, Beidi Chen, and Xia Hu. 2024.
\newblock \href {https://arxiv.org/abs/2402.02750} {{KIVI:} {A} tuning-free asymmetric 2bit quantization for {KV} cache}.
\newblock \emph{ArXiv preprint}.

\bibitem[{Louizos et~al.(2017)Louizos, Welling, and Kingma}]{louizos2017learning}
Christos Louizos, Max Welling, and Diederik~P. Kingma. 2017.
\newblock \href {https://arxiv.org/abs/1712.01312} {Learning sparse neural networks through l\({}_{\mbox{0}}\) regularization}.
\newblock \emph{ArXiv preprint}.

\bibitem[{Lozhkov et~al.(2024)Lozhkov, Ben~Allal, von Werra, and Wolf}]{lozhkov2024fineweb-edu}
Anton Lozhkov, Loubna Ben~Allal, Leandro von Werra, and Thomas Wolf. 2024.
\newblock \href {https://huggingface.co/datasets/HuggingFaceFW/fineweb-edu} {Fineweb-edu}.

\bibitem[{Ma et~al.(2023)Ma, Fang, and Wang}]{ma2023llmPruner}
Xinyin Ma, Gongfan Fang, and Xinchao Wang. 2023.
\newblock \href {http://papers.nips.cc/paper\_files/paper/2023/hash/44956951349095f74492a5471128a7e0-Abstract-Conference.html} {Llm-pruner: On the structural pruning of large language models}.
\newblock In \emph{Advances in Neural Information Processing Systems 36: Annual Conference on Neural Information Processing Systems 2023, NeurIPS 2023, New Orleans, LA, USA, December 10 - 16, 2023}.

\bibitem[{Men et~al.(2024)Men, Xu, Zhang, Wang, Lin, Lu, Han, and Chen}]{shortgpt}
Xin Men, Mingyu Xu, Qingyu Zhang, Bingning Wang, Hongyu Lin, Yaojie Lu, Xianpei Han, and Weipeng Chen. 2024.
\newblock \href {https://arxiv.org/abs/2403.03853} {Shortgpt: Layers in large language models are more redundant than you expect}.
\newblock \emph{ArXiv preprint}.

\bibitem[{Merity et~al.(2017)Merity, Xiong, Bradbury, and Socher}]{wikitext2}
Stephen Merity, Caiming Xiong, James Bradbury, and Richard Socher. 2017.
\newblock \href {https://openreview.net/forum?id=Byj72udxe} {Pointer sentinel mixture models}.
\newblock In \emph{Proc. of ICLR}.

\bibitem[{Meta(2024)}]{llama3}
Meta. 2024.
\newblock \href {https://ai.meta.com/blog/meta-llama-3/} {Introducing meta llama 3: The most capable openly available llm to date}.

\bibitem[{Michel et~al.(2019)Michel, Levy, and Neubig}]{michel2019are}
Paul Michel, Omer Levy, and Graham Neubig. 2019.
\newblock \href {https://proceedings.neurips.cc/paper/2019/hash/2c601ad9d2ff9bc8b282670cdd54f69f-Abstract.html} {Are sixteen heads really better than one?}
\newblock In \emph{Advances in Neural Information Processing Systems 32: Annual Conference on Neural Information Processing Systems 2019, NeurIPS 2019, December 8-14, 2019, Vancouver, BC, Canada}.

\bibitem[{Mihaylov et~al.(2018)Mihaylov, Clark, Khot, and Sabharwal}]{obqa}
Todor Mihaylov, Peter Clark, Tushar Khot, and Ashish Sabharwal. 2018.
\newblock \href {https://aclanthology.org/D18-1260} {Can a suit of armor conduct electricity? a new dataset for open book question answering}.
\newblock In \emph{Proc. of EMNLP}.

\bibitem[{Muralidharan et~al.(2024)Muralidharan, Sreenivas, Joshi, Chochowski, Patwary, Shoeybi, Catanzaro, Kautz, and Molchanov}]{minitron2024}
Saurav Muralidharan, Sharath~Turuvekere Sreenivas, Raviraj Joshi, Marcin Chochowski, Mostofa Patwary, Mohammad Shoeybi, Bryan Catanzaro, Jan Kautz, and Pavlo Molchanov. 2024.
\newblock \href {https://arxiv.org/abs/2407.14679} {Compact language models via pruning and knowledge distillation}.
\newblock \emph{ArXiv preprint}.

\bibitem[{OpenAI(2023)}]{openai2023gpt4}
OpenAI. 2023.
\newblock \href {https://arxiv.org/abs/2303.08774} {{GPT-4} technical report}.
\newblock \emph{ArXiv preprint}.

\bibitem[{Penedo et~al.(2024)Penedo, Kydlíček, allal, Lozhkov, Mitchell, Raffel, Werra, and Wolf}]{penedo2024finewebdatasetsdecantingweb}
Guilherme Penedo, Hynek Kydlíček, Loubna~Ben allal, Anton Lozhkov, Margaret Mitchell, Colin Raffel, Leandro~Von Werra, and Thomas Wolf. 2024.
\newblock \href {https://arxiv.org/abs/2406.17557} {The fineweb datasets: Decanting the web for the finest text data at scale}.

\bibitem[{{Qwen Team}(2024)}]{qwen2.5}
{Qwen Team}. 2024.
\newblock \href {https://qwenlm.github.io/blog/qwen2.5/} {Qwen2.5: A party of foundation models}.

\bibitem[{Sakaguchi et~al.(2020)Sakaguchi, Bras, Bhagavatula, and Choi}]{WinoGrande}
Keisuke Sakaguchi, Ronan~Le Bras, Chandra Bhagavatula, and Yejin Choi. 2020.
\newblock \href {https://aaai.org/ojs/index.php/AAAI/article/view/6399} {Winogrande: An adversarial winograd schema challenge at scale}.
\newblock In \emph{The Thirty-Fourth {AAAI} Conference on Artificial Intelligence, {AAAI} 2020, The Thirty-Second Innovative Applications of Artificial Intelligence Conference, {IAAI} 2020, The Tenth {AAAI} Symposium on Educational Advances in Artificial Intelligence, {EAAI} 2020, New York, NY, USA, February 7-12, 2020}.

\bibitem[{Shao et~al.(2023)Shao, Chen, Zhang, Xu, Zhao, Li, Zhang, Gao, Qiao, and Luo}]{shao2023omniquant}
Wenqi Shao, Mengzhao Chen, Zhaoyang Zhang, Peng Xu, Lirui Zhao, Zhiqian Li, Kaipeng Zhang, Peng Gao, Yu~Qiao, and Ping Luo. 2023.
\newblock \href {https://arxiv.org/abs/2308.13137} {Omniquant: Omnidirectionally calibrated quantization for large language models}.
\newblock \emph{ArXiv preprint}.

\bibitem[{Sheng et~al.(2023)Sheng, Zheng, Yuan, Li, Ryabinin, Chen, Liang, R{\'{e}}, Stoica, and Zhang}]{sheng2023flexgen}
Ying Sheng, Lianmin Zheng, Binhang Yuan, Zhuohan Li, Max Ryabinin, Beidi Chen, Percy Liang, Christopher R{\'{e}}, Ion Stoica, and Ce~Zhang. 2023.
\newblock \href {https://proceedings.mlr.press/v202/sheng23a.html} {Flexgen: High-throughput generative inference of large language models with a single {GPU}}.
\newblock In \emph{International Conference on Machine Learning, {ICML} 2023, 23-29 July 2023, Honolulu, Hawaii, {USA}}, Proceedings of Machine Learning Research.

\bibitem[{Soboleva et~al.(2023)Soboleva, Al-Khateeb, Myers, Steeves, Hestness, and Dey}]{cerebras2023slimpajama}
Daria Soboleva, Faisal Al-Khateeb, Robert Myers, Jacob~R Steeves, Joel Hestness, and Nolan Dey. 2023.
\newblock \href {https://huggingface.co/datasets/cerebras/SlimPajama-627B} {{SlimPajama: A 627B token cleaned and deduplicated version of RedPajama}}.

\bibitem[{Sun et~al.(2024)Sun, Liu, Bair, and Kolter}]{wanda}
Mingjie Sun, Zhuang Liu, Anna Bair, and J.~Zico Kolter. 2024.
\newblock \href {https://openreview.net/forum?id=PxoFut3dWW} {A simple and effective pruning approach for large language models}.
\newblock In \emph{Proc. of ICLR}.

\bibitem[{Taori et~al.(2023)Taori, Gulrajani, Zhang, Dubois, Li, Guestrin, Liang, and Hashimoto}]{alpaca}
Rohan Taori, Ishaan Gulrajani, Tianyi Zhang, Yann Dubois, Xuechen Li, Carlos Guestrin, Percy Liang, and Tatsunori~B. Hashimoto. 2023.
\newblock Stanford alpaca: An instruction-following llama model.
\newblock \url{https://github.com/tatsu-lab/stanford_alpaca}.

\bibitem[{Touvron et~al.(2023{\natexlab{a}})Touvron, Lavril, Izacard, Martinet, Lachaux, Lacroix, Rozi{\`{e}}re, Goyal, Hambro, Azhar, Rodriguez, Joulin, Grave, and Lample}]{llama1}
Hugo Touvron, Thibaut Lavril, Gautier Izacard, Xavier Martinet, Marie{-}Anne Lachaux, Timoth{\'{e}}e Lacroix, Baptiste Rozi{\`{e}}re, Naman Goyal, Eric Hambro, Faisal Azhar, Aur{\'{e}}lien Rodriguez, Armand Joulin, Edouard Grave, and Guillaume Lample. 2023{\natexlab{a}}.
\newblock \href {https://arxiv.org/abs/2302.13971} {Llama: Open and efficient foundation language models}.
\newblock \emph{ArXiv preprint}.

\bibitem[{Touvron et~al.(2023{\natexlab{b}})Touvron, Martin, Stone, Albert, Almahairi, Babaei, Bashlykov, Batra, Bhargava, Bhosale, Bikel, Blecher, Canton{-}Ferrer, Chen, Cucurull, Esiobu, Fernandes, Fu, Fu, Fuller, Gao, Goswami, Goyal, Hartshorn, Hosseini, Hou, Inan, Kardas, Kerkez, Khabsa, Kloumann, Korenev, Koura, Lachaux, Lavril, Lee, Liskovich, Lu, Mao, Martinet, Mihaylov, Mishra, Molybog, Nie, Poulton, Reizenstein, Rungta, Saladi, Schelten, Silva, Smith, Subramanian, Tan, Tang, Taylor, Williams, Kuan, Xu, Yan, Zarov, Zhang, Fan, Kambadur, Narang, Rodriguez, Stojnic, Edunov, and Scialom}]{llama2}
Hugo Touvron, Louis Martin, Kevin Stone, Peter Albert, Amjad Almahairi, Yasmine Babaei, Nikolay Bashlykov, Soumya Batra, Prajjwal Bhargava, Shruti Bhosale, Dan Bikel, Lukas Blecher, Cristian Canton{-}Ferrer, Moya Chen, Guillem Cucurull, David Esiobu, Jude Fernandes, Jeremy Fu, Wenyin Fu, Brian Fuller, Cynthia Gao, Vedanuj Goswami, Naman Goyal, Anthony Hartshorn, Saghar Hosseini, Rui Hou, Hakan Inan, Marcin Kardas, Viktor Kerkez, Madian Khabsa, Isabel Kloumann, Artem Korenev, Punit~Singh Koura, Marie{-}Anne Lachaux, Thibaut Lavril, Jenya Lee, Diana Liskovich, Yinghai Lu, Yuning Mao, Xavier Martinet, Todor Mihaylov, Pushkar Mishra, Igor Molybog, Yixin Nie, Andrew Poulton, Jeremy Reizenstein, Rashi Rungta, Kalyan Saladi, Alan Schelten, Ruan Silva, Eric~Michael Smith, Ranjan Subramanian, Xiaoqing~Ellen Tan, Binh Tang, Ross Taylor, Adina Williams, Jian~Xiang Kuan, Puxin Xu, Zheng Yan, Iliyan Zarov, Yuchen Zhang, Angela Fan, Melanie Kambadur, Sharan Narang, Aur{\'{e}}lien Rodriguez, Robert Stojnic, Sergey Edunov,
  and Thomas Scialom. 2023{\natexlab{b}}.
\newblock \href {https://arxiv.org/abs/2307.09288} {Llama 2: Open foundation and fine-tuned chat models}.
\newblock \emph{ArXiv preprint}.

\bibitem[{Vaswani et~al.(2017)Vaswani, Shazeer, Parmar, Uszkoreit, Jones, Gomez, Kaiser, and Polosukhin}]{vaswani2017attention}
Ashish Vaswani, Noam Shazeer, Niki Parmar, Jakob Uszkoreit, Llion Jones, Aidan~N. Gomez, Lukasz Kaiser, and Illia Polosukhin. 2017.
\newblock \href {https://proceedings.neurips.cc/paper/2017/hash/3f5ee243547dee91fbd053c1c4a845aa-Abstract.html} {Attention is all you need}.
\newblock In \emph{Advances in Neural Information Processing Systems 30: Annual Conference on Neural Information Processing Systems 2017, December 4-9, 2017, Long Beach, CA, {USA}}.

\bibitem[{Voita et~al.(2019)Voita, Talbot, Moiseev, Sennrich, and Titov}]{voita2019analyzing}
Elena Voita, David Talbot, Fedor Moiseev, Rico Sennrich, and Ivan Titov. 2019.
\newblock \href {https://aclanthology.org/P19-1580} {Analyzing multi-head self-attention: Specialized heads do the heavy lifting, the rest can be pruned}.
\newblock In \emph{Proc. of ACL}.

\bibitem[{Wang et~al.(2023)Wang, Wang, Li, Gao, Yin, and Ren}]{wang2023scott}
Peifeng Wang, Zhengyang Wang, Zheng Li, Yifan Gao, Bing Yin, and Xiang Ren. 2023.
\newblock \href {https://doi.org/10.18653/v1/2023.acl-long.304} {{SCOTT:} self-consistent chain-of-thought distillation}.
\newblock In \emph{Proceedings of the 61st Annual Meeting of the Association for Computational Linguistics (Volume 1: Long Papers), {ACL} 2023, Toronto, Canada, July 9-14, 2023}.

\bibitem[{Wang et~al.(2024)Wang, Chen, Zhou, Zhu, Liang, Shan, Liu, Xu, Yang, and Qin}]{SmartTrim}
Zekun Wang, Jingchang Chen, Wangchunshu Zhou, Haichao Zhu, Jiafeng Liang, Liping Shan, Ming Liu, Dongliang Xu, Qing Yang, and Bing Qin. 2024.
\newblock \href {https://aclanthology.org/2024.lrec-main.1300} {Smarttrim: Adaptive tokens and attention pruning for efficient vision-language models}.
\newblock In \emph{Proceedings of the 2024 Joint International Conference on Computational Linguistics, Language Resources and Evaluation, {LREC/COLING} 2024, 20-25 May, 2024, Torino, Italy}.

\bibitem[{Wang et~al.(2022)Wang, Wang, Zhu, Liu, Qin, and Wei}]{wang2022distilled}
Zekun Wang, Wenhui Wang, Haichao Zhu, Ming Liu, Bing Qin, and Furu Wei. 2022.
\newblock \href {https://aclanthology.org/2022.emnlp-main.608} {Distilled dual-encoder model for vision-language understanding}.
\newblock In \emph{Proc. of EMNLP}.

\bibitem[{Wang et~al.(2020)Wang, Wohlwend, and Lei}]{wang2020structured}
Ziheng Wang, Jeremy Wohlwend, and Tao Lei. 2020.
\newblock \href {https://aclanthology.org/2020.emnlp-main.496} {Structured pruning of large language models}.
\newblock In \emph{Proc. of EMNLP}.

\bibitem[{Wolf et~al.(2019)Wolf, Debut, Sanh, Chaumond, Delangue, Moi, Cistac, Rault, Louf, Funtowicz, and Brew}]{hf_transformers}
Thomas Wolf, Lysandre Debut, Victor Sanh, Julien Chaumond, Clement Delangue, Anthony Moi, Pierric Cistac, Tim Rault, R{\'{e}}mi Louf, Morgan Funtowicz, and Jamie Brew. 2019.
\newblock \href {https://arxiv.org/abs/1910.03771} {Huggingface's transformers: State-of-the-art natural language processing}.
\newblock \emph{ArXiv preprint}.

\bibitem[{Xia et~al.(2023)Xia, Gao, Zeng, and Chen}]{ShearedLLaMA}
Mengzhou Xia, Tianyu Gao, Zhiyuan Zeng, and Danqi Chen. 2023.
\newblock \href {https://arxiv.org/abs/2310.06694} {Sheared llama: Accelerating language model pre-training via structured pruning}.
\newblock \emph{ArXiv preprint}.

\bibitem[{Xia et~al.(2022)Xia, Zhong, and Chen}]{xia2022structured}
Mengzhou Xia, Zexuan Zhong, and Danqi Chen. 2022.
\newblock \href {https://aclanthology.org/2022.acl-long.107} {Structured pruning learns compact and accurate models}.
\newblock In \emph{Proc. of ACL}.

\bibitem[{Xiao et~al.(2023)Xiao, Lin, Seznec, Wu, Demouth, and Han}]{Xiao23smoothquant}
Guangxuan Xiao, Ji~Lin, Micka{\"{e}}l Seznec, Hao Wu, Julien Demouth, and Song Han. 2023.
\newblock \href {https://proceedings.mlr.press/v202/xiao23c.html} {Smoothquant: Accurate and efficient post-training quantization for large language models}.
\newblock In \emph{International Conference on Machine Learning, {ICML} 2023, 23-29 July 2023, Honolulu, Hawaii, {USA}}, Proceedings of Machine Learning Research.

\bibitem[{Xu et~al.(2024{\natexlab{a}})Xu, Shao, Chen, Tang, Zhang, Gao, An, Qiao, and Luo}]{xu2024besa}
Peng Xu, Wenqi Shao, Mengzhao Chen, Shitao Tang, Kaipeng Zhang, Peng Gao, Fengwei An, Yu~Qiao, and Ping Luo. 2024{\natexlab{a}}.
\newblock \href {https://arxiv.org/abs/2402.16880} {{BESA:} pruning large language models with blockwise parameter-efficient sparsity allocation}.
\newblock \emph{ArXiv preprint}.

\bibitem[{Xu et~al.(2024{\natexlab{b}})Xu, Han, Yang, Wang, Zhu, Liu, Liu, and Che}]{xu2024onebit}
Yuzhuang Xu, Xu~Han, Zonghan Yang, Shuo Wang, Qingfu Zhu, Zhiyuan Liu, Weidong Liu, and Wanxiang Che. 2024{\natexlab{b}}.
\newblock \href {https://arxiv.org/abs/2402.11295} {Onebit: Towards extremely low-bit large language models}.
\newblock \emph{ArXiv preprint}.

\bibitem[{Yang et~al.(2024{\natexlab{a}})Yang, Yang, Hui, Zheng, Yu, Zhou, Li, Li, Liu, Huang, Dong, Wei, Lin, Tang, Wang, Yang, Tu, Zhang, Ma, Yang, Xu, Zhou, Bai, He, Lin, Dang, Lu, Chen, Yang, Li, Xue, Ni, Zhang, Wang, Peng, Men, Gao, Lin, Wang, Bai, Tan, Zhu, Li, Liu, Ge, Deng, Zhou, Ren, Zhang, Wei, Ren, Liu, Fan, Yao, Zhang, Wan, Chu, Liu, Cui, Zhang, Guo, and Fan}]{qwen2}
An~Yang, Baosong Yang, Binyuan Hui, Bo~Zheng, Bowen Yu, Chang Zhou, Chengpeng Li, Chengyuan Li, Dayiheng Liu, Fei Huang, Guanting Dong, Haoran Wei, Huan Lin, Jialong Tang, Jialin Wang, Jian Yang, Jianhong Tu, Jianwei Zhang, Jianxin Ma, Jianxin Yang, Jin Xu, Jingren Zhou, Jinze Bai, Jinzheng He, Junyang Lin, Kai Dang, Keming Lu, Keqin Chen, Kexin Yang, Mei Li, Mingfeng Xue, Na~Ni, Pei Zhang, Peng Wang, Ru~Peng, Rui Men, Ruize Gao, Runji Lin, Shijie Wang, Shuai Bai, Sinan Tan, Tianhang Zhu, Tianhao Li, Tianyu Liu, Wenbin Ge, Xiaodong Deng, Xiaohuan Zhou, Xingzhang Ren, Xinyu Zhang, Xipin Wei, Xuancheng Ren, Xuejing Liu, Yang Fan, Yang Yao, Yichang Zhang, Yu~Wan, Yunfei Chu, Yuqiong Liu, Zeyu Cui, Zhenru Zhang, Zhifang Guo, and Zhihao Fan. 2024{\natexlab{a}}.
\newblock \href {https://arxiv.org/abs/2407.10671} {Qwen2 technical report}.
\newblock \emph{ArXiv preprint}.

\bibitem[{Yang et~al.(2024{\natexlab{b}})Yang, Cao, and Zhao}]{yang2024laco}
Yifei Yang, Zouying Cao, and Hai Zhao. 2024{\natexlab{b}}.
\newblock \href {https://arxiv.org/abs/2402.11187} {Laco: Large language model pruning via layer collapse}.
\newblock \emph{ArXiv preprint}.

\bibitem[{Yao et~al.(2022)Yao, Aminabadi, Zhang, Wu, Li, and He}]{yao2022zeroquant}
Zhewei Yao, Reza~Yazdani Aminabadi, Minjia Zhang, Xiaoxia Wu, Conglong Li, and Yuxiong He. 2022.
\newblock \href {http://papers.nips.cc/paper\_files/paper/2022/hash/adf7fa39d65e2983d724ff7da57f00ac-Abstract-Conference.html} {Zeroquant: Efficient and affordable post-training quantization for large-scale transformers}.
\newblock In \emph{Advances in Neural Information Processing Systems 35: Annual Conference on Neural Information Processing Systems 2022, NeurIPS 2022, New Orleans, LA, USA, November 28 - December 9, 2022}.

\bibitem[{Yin et~al.(2024)Yin, Wu, Zhang, Hsieh, Wang, Jia, Li, Jaiswal, Pechenizkiy, Liang, Bendersky, Wang, and Liu}]{yin2024owl}
Lu~Yin, You Wu, Zhenyu Zhang, Cheng{-}Yu Hsieh, Yaqing Wang, Yiling Jia, Gen Li, Ajay~Kumar Jaiswal, Mykola Pechenizkiy, Yi~Liang, Michael Bendersky, Zhangyang Wang, and Shiwei Liu. 2024.
\newblock \href {https://openreview.net/forum?id=ahEm3l2P6w} {Outlier weighed layerwise sparsity {(OWL):} {A} missing secret sauce for pruning llms to high sparsity}.
\newblock In \emph{Forty-first International Conference on Machine Learning, {ICML} 2024, Vienna, Austria, July 21-27, 2024}.

\bibitem[{Zellers et~al.(2019)Zellers, Holtzman, Bisk, Farhadi, and Choi}]{HellaSwag}
Rowan Zellers, Ari Holtzman, Yonatan Bisk, Ali Farhadi, and Yejin Choi. 2019.
\newblock \href {https://aclanthology.org/P19-1472} {{H}ella{S}wag: Can a machine really finish your sentence?}
\newblock In \emph{Proc. of ACL}.

\bibitem[{Zhang et~al.(2024{\natexlab{a}})Zhang, Shi, Sun, and Sun}]{zhang2024structured}
Honghe Zhang, Xiaolong Shi, Jingwei Sun, and Guangzhong Sun. 2024{\natexlab{a}}.
\newblock \href {https://doi.org/10.18653/v1/2024.findings-naacl.1} {Structured pruning for large language models using coupled components elimination and minor fine-tuning}.
\newblock In \emph{Findings of the Association for Computational Linguistics: {NAACL} 2024, Mexico City, Mexico, June 16-21, 2024}.

\bibitem[{Zhang et~al.(2023{\natexlab{a}})Zhang, Chen, Shen, Yang, Ou, Yu, and Zhuang}]{loraprune}
Mingyang Zhang, Hao Chen, Chunhua Shen, Zhen Yang, Linlin Ou, Xinyi Yu, and Bohan Zhuang. 2023{\natexlab{a}}.
\newblock \href {https://arxiv.org/abs/2305.18403} {Pruning meets low-rank parameter-efficient fine-tuning}.
\newblock \emph{ArXiv preprint}.

\bibitem[{Zhang et~al.(2024{\natexlab{b}})Zhang, Bai, Lin, Zhao, Hou, and Cannistraci}]{zhang2024plugandplay}
Yingtao Zhang, Haoli Bai, Haokun Lin, Jialin Zhao, Lu~Hou, and Carlo~Vittorio Cannistraci. 2024{\natexlab{b}}.
\newblock \href {https://openreview.net/forum?id=Tr0lPx9woF} {Plug-and-play: An efficient post-training pruning method for large language models}.
\newblock In \emph{The Twelfth International Conference on Learning Representations}.

\bibitem[{Zhang et~al.(2022)Zhang, Lin, Liu, Li, Sun, and Zhou}]{zhang2022moefication}
Zhengyan Zhang, Yankai Lin, Zhiyuan Liu, Peng Li, Maosong Sun, and Jie Zhou. 2022.
\newblock \href {https://aclanthology.org/2022.findings-acl.71} {{M}o{E}fication: Transformer feed-forward layers are mixtures of experts}.
\newblock In \emph{Findings of the Association for Computational Linguistics: ACL 2022}.

\bibitem[{Zhang et~al.(2023{\natexlab{b}})Zhang, Sheng, Zhou, Chen, Zheng, Cai, Song, Tian, R{\'{e}}, Barrett, Wang, and Chen}]{zhang2023h2o}
Zhenyu Zhang, Ying Sheng, Tianyi Zhou, Tianlong Chen, Lianmin Zheng, Ruisi Cai, Zhao Song, Yuandong Tian, Christopher R{\'{e}}, Clark~W. Barrett, Zhangyang Wang, and Beidi Chen. 2023{\natexlab{b}}.
\newblock \href {http://papers.nips.cc/paper\_files/paper/2023/hash/6ceefa7b15572587b78ecfcebb2827f8-Abstract-Conference.html} {{H2O:} heavy-hitter oracle for efficient generative inference of large language models}.
\newblock In \emph{Advances in Neural Information Processing Systems 36: Annual Conference on Neural Information Processing Systems 2023, NeurIPS 2023, New Orleans, LA, USA, December 10 - 16, 2023}.

\bibitem[{Zhu et~al.(2021)Zhu, Wang, Zhang, Liu, Zhao, and Qin}]{zhu2021less}
Haichao Zhu, Zekun Wang, Heng Zhang, Ming Liu, Sendong Zhao, and Bing Qin. 2021.
\newblock \href {https://aclanthology.org/2021.findings-emnlp.95} {Less is more: Domain adaptation with lottery ticket for reading comprehension}.
\newblock In \emph{Findings of the Association for Computational Linguistics: EMNLP 2021}.

\bibitem[{Zuo et~al.(2022)Zuo, Zhang, Liang, He, Zhao, and Chen}]{zuo2022moebert}
Simiao Zuo, Qingru Zhang, Chen Liang, Pengcheng He, Tuo Zhao, and Weizhu Chen. 2022.
\newblock \href {https://aclanthology.org/2022.naacl-main.116} {{M}o{EBERT}: from {BERT} to mixture-of-experts via importance-guided adaptation}.
\newblock In \emph{Proceedings of the 2022 Conference of the North American Chapter of the Association for Computational Linguistics: Human Language Technologies}.

\end{thebibliography}

\appendix
\section{Details of Experimental Setup}

\subsection{Details of Evaluation Benchmarks}
\label{app:benchmarks}

\paragraph{Zero-shot Tasks Evaluation} In this work, we consider the following tasks for evaluating zero-shot performance, along with their respective evaluation metrics: WinoGrande~\citep{WinoGrande} with the accuracy, PIQA~\citep{PIQA} with the normalized accuracy, OBQA~\citep{obqa} with the normalized accuracy, HellaSwag~\citep{HellaSwag} with the normalized accuracy, ARC-easy/challenge~\citep{arc} with the normalized accuracy, MMLU~\citep{mmlu} with the accuracy, FreebaseQA~\citep{freebaseqa} with the exact-match score.
The first 6 tasks are general common sense reasoning tasks, while the others are knowledge-intensive.
We evaluate WinoGrande, PIQA, HellaSwag, BoolQ, ARC-e/c, and MMLU by LM-Evaluation-Harness~\citep{eval-harness} in multiple choice form: we compute the loglikelihood for each choice and report the accuracy for the highest choice.
For FreebaseQA, the evaluation is run with LLM-Kick~\citep{llmkick}.

\paragraph{Language Modeling Evaluation} We evaluate language modeling performance on WikiText2~\citep{wikitext2} validation set with the setting of~\citep{eval-harness}.
The input length is 1024 for the 8B/13B models and 256 for the 70B models.

\paragraph{Inference Efficiency Evaluation}
We evaluate the speed-up of CPU on an Intel
Xeon E5-466 2640 v4 CPU and the speed-up of GPU on a single A800-80G GPU.
We set the sequence length to 1024 and the batch size to 1.

\subsection{Implementation Details}
\label{app:methods_details}
We implement \ours{} with Huggingface Transformer~\citep{hf_transformers}.
We perform experiments on NVIDIA A800-80G GPUs.
The pruning stage is conducted on 1 GPU for the 7B/13B models and 8 GPUs for the 70B models. 
Unless otherwise stated, the calibration dataset consists of 128 samples and each has approximately 1024 tokens following \citet{wanda, flap}. 
By default, for the 7B/8B/13B models, $\alpha$ in Equation~\ref{eq:22} is set to 1, whereas for the 70B model, it is set to 3.
For the recovery fine-tuning stage, the average rank budget is set to 8 by default. 
We explore the following datasets for recovery training:
\begin{itemize}
    \item \textbf{Slimpajama}\footnote{\href{https://huggingface.co/datasets/DKYoon/SlimPajama-6B}{https://huggingface.co/datasets/DKYoon/SlimPajama-6B}}~\citep{cerebras2023slimpajama} is created by cleaning and deduplicating the RedPajama dataset~\citep{together2023redpajama}.
    \item \textbf{Alpaca-Cleaned}\footnote{\href{https://huggingface.co/datasets/yahma/alpaca-cleaned}{https://huggingface.co/datasets/yahma/alpaca-cleaned}} is a cleaned version of the original Alpaca~\citep{alpaca}, which is also used as recovery data in previous LLM structural pruning work~\citet{ma2023llmPruner, ashkboos2024slicegpt}.
    \item \textbf{Knowledge-Pile}\footnote{\href{https://huggingface.co/datasets/Query-of-CC/Knowledge_Pile}{https://huggingface.co/datasets/Query-of-CC/Knowledge\_Pile}}~\cite{fei2024query} is a dataset with high-quality knowledge data retrieved from public corpora.
    \item \textbf{FineWeb-Edu}\footnote{\href{https://huggingface.co/datasets/HuggingFaceFW/fineweb-edu}{https://huggingface.co/datasets/HuggingFaceFW/fineweb-edu}} is a dataset filtered from FineWeb~\citep{penedo2024finewebdatasetsdecantingweb}, focusing on high-quality educational web pages using a classifier trained with annotations from LLaMA3-70B-Instruct.
\end{itemize}
In our experiments, we use FinWeb-Edu as our default recovery data since it achieves the best performance across all tasks.
More experimental results are shown in Appendix~\ref{app:more_analysis}.
We use the AdamW optimizer with a learning rate of 2e-4 for the 8B and 13B models, and 1e-4 for the 70B models.
The batch size is set to 128.
We use 8 GPUs to fine-tune the pruned 7B/8B/13B models and 16 GPUs for the 70G models. We also show the details of time cost for pruning and recovery in Table~\ref{tab:time}.
\begin{table}[t!]
\centering
\resizebox{1.0\linewidth}{!}{

\begin{tabular}{c|cc|cc}
\hline
\multirow{2}{*}{\textbf{Model Size}} & \multicolumn{2}{c|}{\textbf{prune}} & \multicolumn{2}{c}{\textbf{recovery}} \\
\cline{2-5}
& \textbf{device} & \textbf{time} & \textbf{device} & \textbf{time} \\
\hline
7/8B & 1xA800-80G & 2min & 8xA800-80G & 0.5h \\
13B & 1xA800-80G & 4-5min & 8xA800-80G & 0.92h \\
70B & 8xA800-80G & 15min & 16xA800-80G & 5.42h \\
\hline
\end{tabular}

}

\caption{Details of time cost of pruning and recovery for different model sizes.}
\label{tab:time}
\end{table}

\subsection{Details of Baselines}
\label{app:baseline_details}
In this section, we present more details of baselines in comparison:
\paragraph{Magnitude-SP} measures the importance criterion based on the magnitude of weights~\citep{han2015deep,jaiswal2023theemergence}. This baseline employs with uniform sparsity across blocks.

\paragraph{Wanda-SP} is extended by the unstructured pruning method Wanda~\citep{wanda}, which modifies the target pruning units to structured weights.
This baseline uses the product of weights and activations as an importance criterion. 
We globally sort the pruning units across all blocks to identify redundant components, as this strategy tends to achieve better performance compared to adopting a local manner for individual blocks.

\paragraph{FLAP}\citep{flap} is a training-free structured pruning method for LLM, using the stability of activations as an importance criterion with a global sorting strategy.
We follow its optimal setting: \textit{Weighted Input Feature Variance}.

For a fair comparison, all baselines are implemented to prune the intermediate dimensions of FFN, which are the same as \ours{}. Since the original FLAP paper only reports the results of both MHA and FFN pruning on LLaMA, we reimplement based on their official code and conduct on more models.

\begin{table}[t!]
\small
  \centering
  \setlength{\tabcolsep}{4.5pt}
  \begin{tabular}{l|l|cc}
  \toprule
  Sparsity & Method  & Average & PPL \\
  \midrule
    0\%  & Qwen2.5-7B & 68.58 & 7.64 \\
    \midrule
    \multirow{4}{*}{20\%} & \multicolumn{2}{l}{\textit{w/o recovery}} \\
   & Wanda-SP &  62.70 & \bf 8.82 \\
    & FLAP &  61.71 & 9.12 \\
   & \ours{} (ours) &  \bf 63.02 &  9.03\\
  \bottomrule
  \end{tabular}
  \caption{The averaged zero-shot performance and PPL on wikitext2 of pruned models on Qwen2.5-7B under 20\% sparsity. 
  \textbf{Bold} indicates the best results.}
  \label{tab:zs_qwen25_7b}
\end{table}
\begin{table*}[t!]
  \centering
  \setlength{\tabcolsep}{4.5pt}
  \resizebox{1.0\linewidth}{!}{
  \begin{tabular}{l|l|ccccccc|c}
  \toprule
 Sparsity & Method &   WinoGrande & PIQA
 & OBQA & HellaSwag & ARC-e & ARC-c & MMLU & Average \\
  \midrule
    0\% & LLaMA3.1-8B & 73.95 & 81.01 & 44.8 & 78.91 & 80.85 & 53.33 & 62.95 & 69.74\\
    \midrule
    \multirow{4}{*}{20\%} & \multicolumn{8}{l}{\textit{w/o recovery}}  \\
    & Wanda-SP & 67.96 & 74.65 & 41.00 & 65.95 &  64.44 & 39.85 &45.21 & 57.01\\
   & FLAP & 64.64 & 73.72 & 40.60 & 61.92 &  61.11 & 35.92 & 38.20 & 53.73\\
    & \ours{} (ours) & \bf71.51 & \bf76.88 & \bf 41.60 & \bf72.28 &  \bf70.39 & \bf44.88 &\bf 54.59 & \bf63.59\\
    \midrule
    \multirow{7}{*}{50\%} & \multicolumn{8}{l}{\textit{w/o recovery}}  \\
    & Wanda-SP & 58.88 & 63.76 & 32.20 &46.03 &  46.93 & 29.52 &26.39 &43.39  \\ 
    & FLAP & 58.09 & 59.96 & 31.00 & 41.98 &  39.56 & 26.02 & 23.15 &  39.97\\ 
    & \ours{} (ours) & \bf{61.09} & \bf{66.16} & \bf{32.40} & \bf{49.31} & \bf{48.70} & \bf{29.95} & \bf{32.05} & \bf{45.67} \\
    \cmidrule(lr){2-10} 
    & \multicolumn{8}{l}{\textit{w/ recovery}} \\
    & Wanda-SP & 61.88 & 70.78 & \bf 36.80 & 59.58 & 61.53 & 36.43 &36.44  & 51.92 \\ 
    & \ours{} & \bf{65.19} & \bf 71.16 & 36.40& \bf 61.23 &  \bf 62.54 & \bf 37.29 & \bf40.65 & \bf{55.83} \\ 
  \bottomrule
  \end{tabular}
  }
    \caption{Zero-shot performance of pruned models on LLaMA3.1-8B under 20\% and 50\% sparsity.
  For 50\% sparsity, we also show the results after recovery fine-tuning.
  \textbf{Bold} results indicate the best results under the same setting.}
  \label{tab:zs_llama31_8b}
\end{table*}

\begin{table*}[t!]
  \centering
  \setlength{\tabcolsep}{4.5pt}
  \resizebox{1.0\linewidth}{!}{
  \begin{tabular}{l|l|ccccccc|c}
  \toprule
 Sparsity & Method &   WinoGrande & PIQA
 & OBQA & HellaSwag & ARC-e & ARC-c & MMLU & Average \\
  \midrule
    0\% & LLaMA-7B & 70.09 & 79.16 & 44.06 & 76.21 & 72.85 & 44.80 & 29.92 & 59.58\\
    \midrule
    \multirow{5}{*}{20\%} & \multicolumn{8}{l}{\textit{w/o recovery}}  \\
     & Magnitude-SP & 49.33 & 52.12 & 24.20 & 27.20 & 28.66 & 25.68 & 24.85 & 33.15\\
    & Wanda-SP & 67.88 & \bf76.17 & 41.00 & 70.54 & 66.67 & 39.85 & 27.63 & 55.68\\
   & FLAP & 66.61 & 75.63 & \bf42.00 & 68.91 & 66.33 & 38.82 & 26.95 & 55.04\\
    & \ours{} (ours) & \bf68.43 & 75.90 & 41.20 & \bf71.48 & \bf69.44 & \bf42.66 & \bf 27.75 & \bf56.69\\
    \midrule
    \multirow{8}{*}{50\%} & \multicolumn{8}{l}{\textit{w/o recovery}}  \\
     & Magnitude-SP & 52.01 & 49.24 & 26.20 & 26.31 & 26.43 & 26.96 & 24.87 & 33.15 \\   
    & Wanda-SP & 63.30 & 65.38 & 37.00 & 52.13 & 47.81 & 29.10 & 24.16 & 45.55  \\ 
    & FLAP & 60.14 & 65.56 & 36.00 & 50.23 & 44.82 & 29.01 & 24.46 & 44.32 \\ 
    & \ours{} (ours) & \bf{63.69} & \bf{66.21} & \bf{37.20} & \bf{54.55} & \bf{47.98} & \bf{30.12} & 24.03 & \bf{46.25} \\
    \cmidrule(lr){2-10} 
    & \multicolumn{8}{l}{\textit{w/ recovery}} \\
    & Wanda-SP & 65.51 & \bf 71.33 & 38.20 & 61.29 & 58.42 & 34.04 & 24.53 & 50.47 \\ 
    & \ours{} & \bf{65.55} & 71.22 & \bf 39.20 & \bf 61.31 & \bf 58.96 & \bf 34.73 & 25.35 & \bf{50.90} \\ 
  \bottomrule
  \end{tabular}
  }
  \caption{Zero-shot performance of pruned models on LLaMA-7B under 20\% and 50\% sparsity.
  For 50\% sparsity, we also show the results after recovery fine-tuning.
  \textbf{Bold} indicates the best results under the same setting.}
  \label{tab:zs_llama_7b}
\end{table*}

\begin{table*}[t!]
  \centering
  \setlength{\tabcolsep}{4.5pt}
  \resizebox{1.0\linewidth}{!}{
  \begin{tabular}{l|l|ccccccc|c}
  \toprule
  Sparsity & Method  & WinoGrande & PIQA
 & OBQA & HellaSwag & ARC-e & ARC-c & MMLU & Average \\
  \midrule
    0\% & LLaMA-13B   & 72.69 & 80.20 & 44.80 &  79.08  &  74.71 & 47.70 & 41.24 & 62.92 \\
    \midrule
    \multirow{5}{*}{20\%} & \multicolumn{8}{l}{\textit{w/o recovery}}  \\
    & Magnitude-SP &   48.86 & 58.43 & 27.40 & 33.29 & 33.80 & 29.10 & 23.36 & 36.32 \\
    &Wanda-SP & 70.56 & 77.53 & 41.40 & 75.40 & 66.25 & 41.98 & 31.65 & 57.82 \\
    & FLAP & 70.56  & 77.09 & 41.40 & 74.19 & 68.77 & 43.60 & 31.77 & 58.19 \\
    & \ours{} (ours) & \bf{71.43} & \bf 78.13 & \bf 42.80 & \bf 76.31 & \bf 68.81 & \bf 45.14 & \bf 38.07 & \bf 60.10 \\
    \midrule
    \multirow{8}{*}{50$\%$} & \multicolumn{8}{l}{\textit{w/o recovery}}\\
    & Magnitude-SP & 50.99 & 50.98 & 26.20 & 27.16 & 27.27 & 27.30 & 23.94 & 33.40 \\   
    & Wanda-SP & 67.01 & 67.46 & 37.00 & 61.44  & 49.83 & 31.14 & 27.23 & 48.73 \\ 
    & FLAP & 64.24 & 70.24 & 36.00 & 56.73 & 52.86 & 33.19 & 25.23 & 48.35 \\ 
      & \ours{} (ours) & \bf 68.43 & \bf 71.76 & \bf 37.60 & \bf 63.50 & \bf 59.85 & \bf 37.46 & \bf 27.38 & \bf 52.28 \\ 
      \cmidrule(lr){2-10} 
    & \multicolumn{8}{l}{\textit{w/ recovery}} \\
    & Wanda-SP & 67.48 & \bf 75.08 & 39.20 & 68.14 & 39.59 & 64.48 & 31.97 & 55.13 \\ 
    & \ours{} (ours) &  \bf 68.82 & 74.76 &  \bf 41.40 & \bf 68.85 & \bf 41.30 & \bf67.13 & \bf 35.63 & \bf 56.84 \\ 
  \bottomrule
  \end{tabular}}
  \caption{Zero-shot performance of pruned models on LLaMA-13B under 20\% and 50\% sparsity.
  For 50\% sparsity, we also show the results after recovery fine-tuning.
  \textbf{Bold} indicates the best results under the same setting.}
  \label{tab:zs_llama_13b}
\end{table*}

\begin{table*}[t!]
  \centering
  \setlength{\tabcolsep}{4.5pt}
  \resizebox{1.0\linewidth}{!}{
  \begin{tabular}{l|l|cccccccc|c}
  \toprule
  Sparsity & Method  & WinoGrande & PIQA
 &  OBQA & HellaSwag & ARC-e & ARC-c & MMLU & FreebaseQA & Average \\
  \midrule
    0\%  & LLaMA2-7B & 69.06 & 79.11 & 44.20 & 76.02 & 74.62 & 46.33 & 41.25 & 68.39 & 62.37 \\
    \midrule
    \multirow{5}{*}{20\%} & \multicolumn{8}{l}{\textit{w/o recovery}} \\
    & Magnitude-SP &  48.70 & 52.12 & 24.40 & 28.77 & 30.18 & 24.32 & 25.84 & 0.55 & 29.36 \\
   & Wanda-SP & 66.93 & 76.50 & \bf 41.80 & 70.82 & 64.48 & 38.57 & 32.19 & \bf 44.44 & 54.46 \\
    & FLAP & 65.51 & 75.84 & 40.00 & 69.64 & 60.82 & 37.29 & 30.86 & 28.65 & 51.07 \\
   & \ours{} (ours) & \bf 67.25 & \bf 76.88 & 40.60 & \bf 72.05 & \bf 68.77 & \bf 41.47 & \bf 36.33 & 38.99 & \bf 55.29\\
    \midrule
    \multirow{8}{*}{50$\%$} & \multicolumn{8}{l}{\textit{w/o recovery}}\\
    & Magnitude-SP & 50.20 & 48.20 & 27.00 & 26.32 & 26.52 & 29.10 & 26.84 & 0.53 & 29.34 \\ 
     & Wanda-SP & 61.56 & 66.49 &  34.80 & 52.23& 45.37 & 28.07 & \bf 25.45 & 4.15 & 39.76 \\ 
     & FLAP & 57.62 & 66.00 & 32.80 & 48.09 & 40.07 & 27.39 & 23.04 & 0.90 & 36.99 \\ 
     & \ours{} (ours) & \bf 61.64 & \bf 67.36 & \bf35.20 & \bf53.96 & \bf48.61 & \bf30.20 & 23.07 & \bf 4.35 & \bf 40.55 \\
      \cmidrule(lr){2-11} 
    & \multicolumn{8}{l}{\textit{w/ recovery}} \\
    &  Wanda-SP & 63.85 & \bf 71.11 & 37.80 & 61.40 & 57.08 & 35.04 & 26.32 & \bf 20.90 & 46.69 \\ 
     & \ours{} (ours) & \bf 65.11  & 70.73 & \bf 37.00 & \bf 61.96 & \bf 58.88 & \bf 36.26 & \bf 29.46 & 20.05 & \bf 47.43  \\ 
  \bottomrule
  \end{tabular}
      }
  \caption{Zero-shot performance of pruned models on LLaMA2-7B under 20\% and 50\% sparsity.
  For 50\% sparsity, we also show the results after recovery fine-tuning. 
  \textbf{Bold} indicates the best results under the same setting.}
  \label{tab:zs_llama2_7b}
\end{table*}

\section{More Results and Analysis}

\subsection{More Results on LLaMA}
\label{app:more_llama_results}
\paragraph{Results of LLaMA3.1-8B}
In addition to the experiments presented in Section~\ref{subsec:main_results}, we also conduct experiments on the latest powerful model LLaMA3.1-8B.

The results are shown in Table~\ref{tab:zs_llama31_8b}.
In the single-shot pruning setting, \ours{} consistently outperforms other baselines across a variety of tasks and sparsity budgets.
Furthermore, we experiment with recovery fine-tuning for \ours{} and Wanda-SP. As observed with previous models, our method still achieves better results.

\paragraph{Results of LLaMA2-7B}
Table~\ref{tab:zs_llama2_7b} shows the results on LLaMA2-7B. In the single-shot pruning setting, \ours{} consistently exceeds other baselines on various tasks and sparsity levels. Additionally, we perform recovery fine-tuning for both \ours{} and Wanda-SP. As with other models, \ours{} also provides superior performance.

\paragraph{Results of LLaMA1}

Since the LLaMA1 family models were released earlier and no longer the best open-source LLMs, we do not include their results in Section~\ref{subsec:main_results}. 
Here, we present the zero-shot performance comparison of LLaMA1 family in Table~\ref{tab:zs_llama_7b} and Table~\ref{tab:zs_llama_13b}\footnote{
The results of Wanda-SP reported by us differ from those in \citet{flap} since we employ a global sorting strategy as described in Appendix~\ref{app:baseline_details}.}. 
It can be observed that on LLaMA-7B and LLaMA-13B, \ours{} consistently achieves the best average performance at different sparsity. 
An interesting phenomenon is that on some challenging tasks (\textit{e.g.} MMLU), all pruning methods exhibit performance close to chance-level accuracy at 50\% sparsity. 
Compared to the results on the LLaMA3 herd of models, this could be attributed to the LLaMA1 family models' inherently weaker performance on these tasks, with high-sparsity pruning further degrading this aspect of their performance. 

\paragraph{Results of Qwen2.5}

In addition to the models of the LLaMA families, we also conduct experiments on Qwen2.5-7B~\citep{qwen2.5} to verify whether our pruning framework is model-agnostic. 
As shown in Table~\ref{tab:zs_qwen25_7b}, \ours{} consistently shows better zero-shot performance on average and achieves comparable PPL.

\begin{figure}[t!]
\centering
\includegraphics[width=0.9\linewidth]{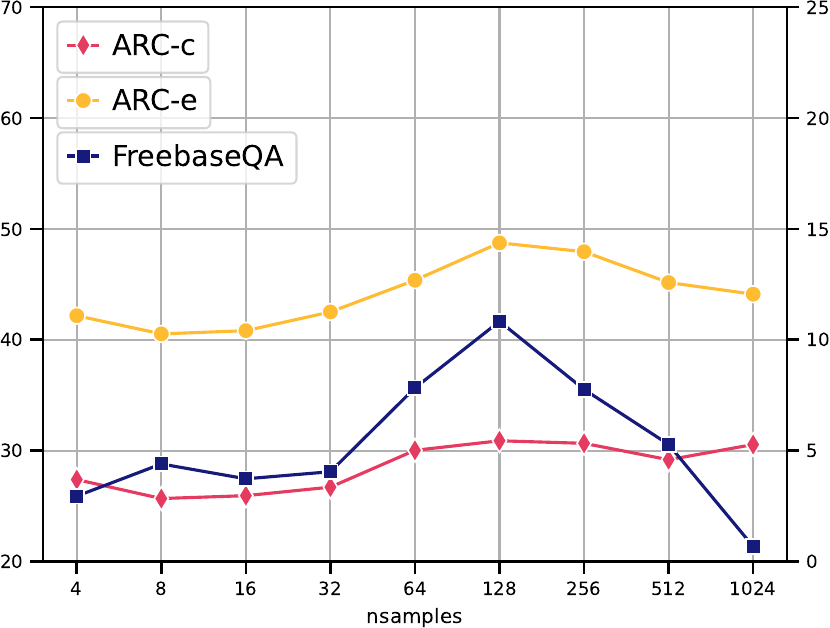}
\caption{
The impact of the size of calibration data. The models are pruned from LLaMA3-8B under 50\% sparsity.
}\label{fig:fig_samples}
\end{figure}

\subsection{More Analysis of \ours{}}
\label{app:more_analysis}

\begin{figure}[t!]
\centering
\includegraphics[width=0.9\linewidth]{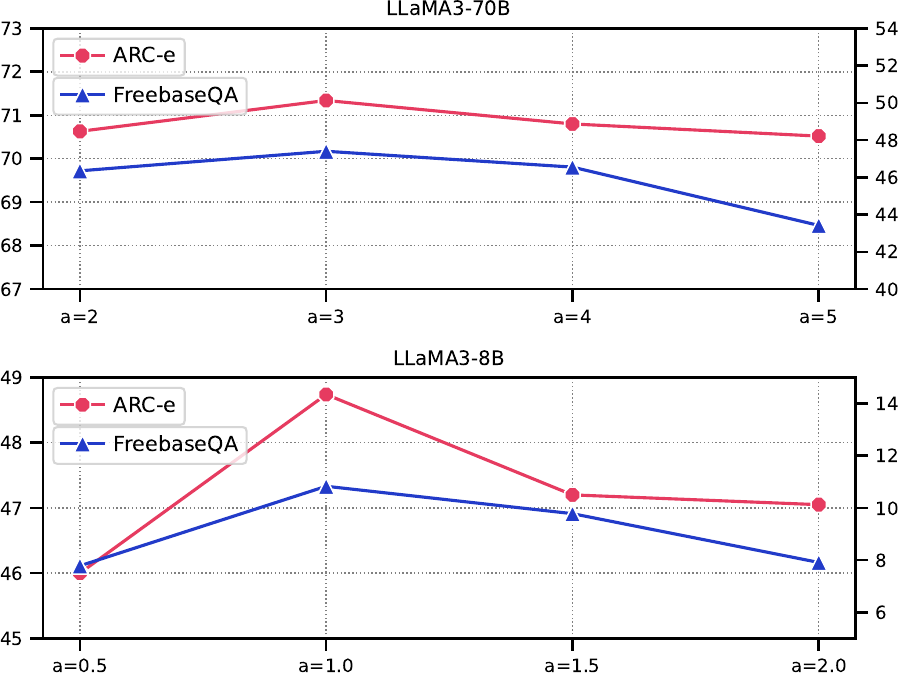}
\caption{
The effect of hyperparameter $\mathbf{ \alpha}$ in calculating block importance. The models are pruned under 50\% sparsity.
}\label{fig:fig_a_best}
\end{figure}

\begin{table*}[t!]
  \centering
  \setlength{\tabcolsep}{4.5pt}
  \resizebox{1.0\linewidth}{!}{
  \begin{tabular}{l|l|ccccccc|c}
  \toprule
 Model & Datasets &   WinoGrande & PIQA
 & OBQA & HellaSwag & ARC-e & ARC-c & MMLU & Average \\
 \midrule
    \multirow{4}{*}{LLaMA2-7B} & Slimpajama & 64.01 & 70.40 & 36.40 & 60.65 & 56.78 & 33.36 & 24.95 & 47.46\\
    & Alpaca-cleaned& 64.33 & 70.24 & \bf 37.60 & 61.22 & \bf 59.30 & 34.47 & 24.33 & 50.21\\
   & Knowledge-pile& 64.80 & 70.13 & 36.80 & 60.45 & 58.42 & 34.81 & 24.55 & 49.99\\
   & FineWeb-edu  & \bf 65.11 & \bf 70.73 & 37.00 & \bf 61.96 & 58.88 & \bf 36.26 & \bf 29.46 & \bf 51.34\\
    \midrule
    \multirow{4}{*}{LLaMA3-8B} & Slimpajama & 64.17 & 70.95 & 35.00 & 60.62 & 57.03 & 33.62 & 37.55 & 51.28\\
    & Alpaca-cleaned& 59.67& 67.85 & 34.80 & 60.97 & 57.41 & 35.24 & 37.89 & 50.55\\
   & Knowledge-pile& \bf 66.14 & 71.38 & 35.60 & 60.89 & 62.05 & 36.69 & 38.07 & 52.97\\
   & FineWeb-edu  & 65.51 & \bf 72.03 & \bf 36.20 & \bf 61.45 & \bf 62.37 & \bf 37.54 & \bf 40.37 & \bf 53.64\\

  \bottomrule
  \end{tabular}
  }
  \caption{Zero-shot performance of various datasets for recovery fine-tuning. All methods are trained with the same tokens (0.1B).
  \textbf{Bold} indicates the best results on each model.}
  \label{tab:datasets}
\end{table*}

\paragraph{Impact of Hyperparameter $\mathbf{ \alpha}$}
In Equation~\ref{eq:22}, we introduce a hyperparameter $\alpha$ to control the intensity of significance during calculating block importance. 
In preliminary experiments, we explore the impact of different $\alpha$ and the results are shown in Figure~\ref{fig:fig_a_best}. 
We observe that for smaller models like LLaMA3-8B, a smaller $\alpha$ is better, while for larger models like the LLaMA3-70B model, a larger $\mathbf{ \alpha}$ is more appropriate.
Finally, we set $\alpha = 1$ for the 7B/8B/13B models and $\alpha=3$ for the 70B models.

\paragraph{Impact of Calibration Data Sizes}

We investigate the impact of calibration data sizes. Figure~\ref{fig:fig_samples} presents the results of 3 tasks on LLaMA3-8B with 20\% sparsity. 
We find that the data with 128 examples yield the best overall performance.

\paragraph{Impact of Recovery Data}

As described in Appendix~\ref{app:methods_details}, we explore various datasets for recovery fine-tuning. 
As shown in Table~\ref{tab:datasets}, FineWeb-Edu consistently outperforms others in a variety of tasks, particularly demonstrating significant improvements in knowledge-intensive tasks such as MMLU and FreebaseQA, which is shown challenging for pruned models~\citep{llmkick}.
Thus, we select it for recovery fine-tuning.

\end{document}